\documentclass[preprint]{jmlr}

\usepackage{longtable}
\usepackage{notation_neureps}
\usepackage{booktabs}
\usepackage[load-configurations=version-1]{siunitx}
\usepackage{subcaption}
\usepackage{tikz}
\usetikzlibrary{positioning,arrows.meta}

\graphicspath{{figs_updated/}{figs/}}

\makeatletter
\renewcommand\ps@jmlrtps{%
  \let\@mkboth\@gobbletwo
  \def\@oddhead{\scriptsize Preprint\hfill}%
  \let\@evenhead\@oddhead
  \def\@oddfoot{}%
  \let\@evenfoot\@oddfoot}
\renewcommand*{\@jmlrenddoc}{%
  \phantomsection
  \protected@edef\@currentlabelname{end of \@shorttitle}%
  \label{jmlrend}%
  \global\let\@reprint\@empty}
\makeatother

\newcommand{\pv}{\bm{p}}
\definecolor{cfull}{HTML}{457B9D}
\newcommand{\hsrc}{\bm h_{\mathrm{src}}}

\theorembodyfont{\upshape}
\theoremheaderfont{\scshape}
\theorempostheader{:}
\theoremsep{\newline}

\jmlrvolume{}
\firstpageno{1}
\jmlryear{2026}

\title[Order and Identity in Number Manifolds]{On the Dissociation of Order and Identity in Number Manifold Representations}

 \author{\Name{Sai Sumedh R. Hindupur} \\
  \addr Harvard John A. Paulson School of Engineering and Applied Sciences
  \AND
  \Name{Thomas Fel} \\
 \addr Goodfire AI
 \AND
    \Name{Demba Ba} \\
 \addr Harvard John A. Paulson School of Engineering and Applied Sciences, \\
 Kempner Institute, Harvard University
 }

\begin{document}

\maketitle

\begin{abstract}
Low-dimensional manifolds have been widely observed in neural network representations, even for ordered concepts such as numbers, whose values admit a natural linear ordering. Recent findings on the efficacy of manifold steering and geometric transformations in computation motivate a \emph{crystal-like} perspective, wherein the manifold constitutes a coherent computational object whose geometry must be respected. However, established findings on the causal efficacy of linear representations challenge this perspective: why are interventions that disregard manifold geometry effective in certain cases?
To reconcile these observations, we propose a \emph{clay-like} perspective on manifold geometry, wherein distinct geometric components of a manifold serve different computational roles: a projection onto a line may support one operation, while the orthogonal component supports another. We demonstrate this in a minimal attention model trained to compute running maxima, a task comprising comparison and identity readout. We show that the norm-minimizing representation for comparison under prescribed margins is a positive rank-one ray, from which a bias-free linear decoder cannot recover multiple identities. We derive sufficient conditions for a functional decomposition into an ordered linear component (the order component) supporting comparison and its orthogonal complement (the readout component) supporting identity readout. The trained model consistently realizes this decomposition, and interventions on the two components selectively steer the corresponding operations.
Our findings reconcile the computational utility of linear representations with nonlinear manifold geometry for ordered concepts. More broadly, understanding how models compute with manifolds requires more than characterizing their geometry: it requires identifying the computational roles of their geometric components.\footnote{Code to reproduce results: \url{https://github.com/Sai-Sumedh/running-max-geometry}.}
\end{abstract}

\section{Introduction}
\label{sec:intro}

The linear representation hypothesis (LRH) posits that concepts are encoded along directions in activation space \citep{park2024linearrepresentationhypothesisgeometry}, a perspective supported by the causal efficacy of interventions along such directions \citep{marks2024geometrytruthemergentlinear,tigges2023linearrepresentationssentimentlarge,voynov2020unsuperviseddiscoveryinterpretabledirections,lee2024mechanisticunderstandingalignmentalgorithms,zhu2024languagemodelsrepresentbeliefs}. However, a growing body of work has identified nonlinear, low-dimensional manifold representations \citep{engels2025languagemodelfeaturesonedimensionally,bhalla2026sparseautoencoderscaptureconcept,fel2026structuringsparsityblocksparsefeaturizers,gurnee2026modelsmanipulatemanifoldsgeometry}, even for ordered concepts such as numbers, whose values admit a natural linear ordering \citep{kantamneni2025languagemodelsusetrigonometry,hindupur2026comparison}. Although theoretical accounts have attributed the emergence of manifold geometry to symmetries in the data or task \citep{karkada2026symmetry,hwang2026intrinsic}, or to representational packing constraints \citep{gurnee2026modelsmanipulatemanifoldsgeometry}, explaining why a representation acquires a particular geometry does not establish how that geometry is used in computation.

Recent studies have begun to establish the causal relevance of manifold geometry through successful manifold steering \citep{wurgaft2026manifold} and the identification of computations that operate through geometric transformations of manifolds \citep{gurnee2026modelsmanipulatemanifoldsgeometry}. These findings motivate what we term a \emph{crystal-like perspective}: the manifold constitutes a coherent computational object whose geometry must be respected. Nonetheless, \citet{hindupur2026comparison} demonstrate that language models compare numbers using linear directions despite representing them on curved manifolds. Similarly, \citet{gross2024compact} identify a dominant key-side size direction in the attention-score matrix of transformers trained on Max-of-$K$. If linear representations suffice for comparison, what computational role does the remaining manifold geometry serve? This apparent discrepancy has important implications for mechanistic interpretability: assuming that computation relies on the entirety of an observed geometry may lead to incorrect characterizations of computational mechanisms and obscure opportunities for selective interventions.

To reconcile these observations, we propose a \emph{clay-like perspective}, wherein distinct geometric components of a manifold serve different computational roles. For ordered concepts, flattening a manifold onto a line through linear projection may support operations that depend on order, while the orthogonal component may support operations that require distinguishing individual concept values.

We demonstrate this perspective in a minimal attention model trained to compute the running maximum of a sequence of numbers (Fig.~\ref{fig:model}). With identity query, key, and value matrices, the model must learn representations that simultaneously satisfy the geometric requirements of two operations: selecting the maximum (\emph{comparison}) and recovering its identity (\emph{identity readout}).

\paragraph{Contributions.}
\begin{enumerate}
    \item \textbf{Theoretical characterization (Sec.~\ref{sec:theory}).} We establish that the norm-minimizing representation for comparison under prescribed margins is a positive rank-one ray, from which a bias-free linear decoder cannot recover multiple identities. We derive sufficient conditions for a functional decomposition into an ordered linear component (the \emph{order component}) supporting comparison and its orthogonal complement (the \emph{readout component}) supporting identity readout.
    \item \textbf{Empirical characterization (Sec.~\ref{sec:geometry}).} We demonstrate the emergence of this decomposition across four training seeds and show that the geometric comparison margin predicts model errors, identifying regimes of interference between the two components.
    \item \textbf{Causal validation (Sec.~\ref{sec:causal}).} We demonstrate that interventions on the two components selectively steer comparison and identity readout without preserving the original manifold geometry.
\end{enumerate}

\section{A Minimal Model of Comparison}
\label{sec:model}

\begin{figure}[t]
\centering
\begin{tikzpicture}[
  font=\small,
  node distance=8mm,
  blk/.style={draw, rounded corners=2pt, align=center, minimum height=9mm, inner sep=3pt, line width=0.6pt},
  trained/.style={blk, fill=cfull!15, draw=cfull!85!black},
  fixed/.style={blk, fill=black!6, draw=black!50},
  note/.style={font=\footnotesize, text=black!70, align=center},
  arr/.style={-{Stealth[length=2.2mm]}, line width=0.7pt}
]
\node[align=center] (in) {$x_1,\dots,x_t$};
\node[trained, right=of in] (emb) {Embedding $\E$\\ (trained)};
\node[fixed, right=10mm of emb] (att) {Causal self-attention\\ $W_Q,W_K,W_V$ fixed to $I$};
\node[trained, right=10mm of att] (dec) {Linear readout $\U$\\ no bias (trained)};
\node[align=center, right=of dec] (out) {$\hat y_t$};
\draw[arr] (in) -- (emb);
\draw[arr] (emb) -- node[above, note]{$\bm h_{t'}$} (att);
\draw[arr] (att) -- node[above, note]{$\bm z_t$} (dec);
\draw[arr] (dec) -- (out);
\node[note, below=2mm of emb] {$\bm h_{t'}=\E_{x_{t'}}$};
\node[note, below=2mm of att] {scores $\bm h_t\cdot\bm h_{t'}/\sqrt d$\\ \textit{comparison}: select maximum};
\node[note, below=2mm of dec] {logits $\U_j\cdot\bm z_t$\\ \textit{identity}: decode value};
\end{tikzpicture}
\caption{\textbf{The running-maximum model: attention scores are inner products of trained embeddings.} Blue denotes trainable parameters; grey denotes fixed operations.}
\label{fig:model}
\end{figure}
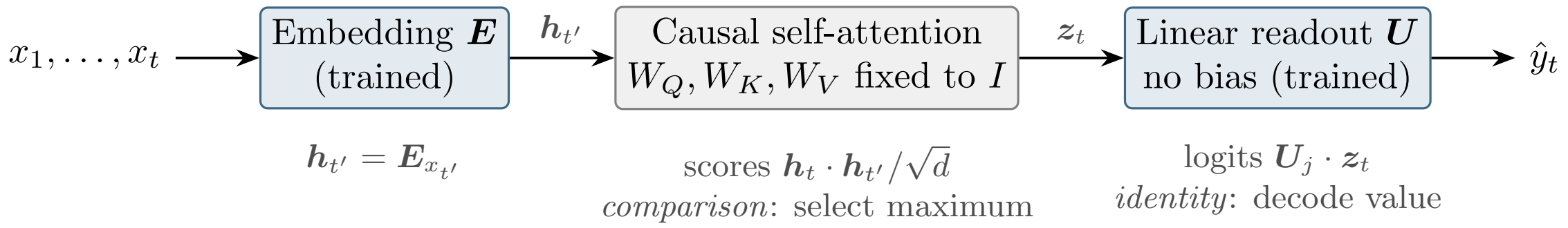

\paragraph{Task and model.} A sequence $x_1,\dots,x_T$ is drawn i.i.d.\ and uniformly from the numbers $\{1,\dots,V\}$, and the target at every position $t$ is the maximum of the prefix, $y_t=\max(x_1,\dots,x_t)$. The model embeds the number $x_t$ as $\bm h_t=\E_{x_t}$ with a trained table $\E\in\R^{V\times d}$, combines these embeddings with one causal self-attention layer whose query, key and value maps are fixed to identity matrices, $\bm z_t=\sum_{t'\le t}\alpha_{tt'}\bm h_{t'}$ with $\alpha_{tt'}=\operatorname{softmax}_{t'\le t}(\bm h_t\cdot\bm h_{t'}/\sqrt d)$, and reads out the logit $\U_j\cdot\bm z_t$ for each number $j$ with a trained, bias-free table $\U\in\R^{V\times d}$ (Fig.~\ref{fig:model}). With $V=1000$, $T=100$ and $d=128$, the model achieves over 99.9\% token-level validation accuracy across all four training seeds (App.~Table~\ref{tab:seeds}).

\paragraph{Notation.} We write $\E_i$ and $\U_i$ for the embedding and unembedding of number $i$. Within one comparison, $q$ is the number at the query position, $m$ is the running maximum and $k$ is any other key, so that $k<m$; since the query is itself a key, $q$ can coincide with $m$ or $k$. To distinguish the geometric contributions to comparison and identity readout, we decompose each vector along a unit direction $\pv$, the \textit{order direction}: $\E_i=a_i\pv+\r_i$ and $\U_i=b_i\pv+\s_i$, with $a_i=\E_i\cdot\pv$, $b_i=\U_i\cdot\pv$ and $\r_i,\s_i\perp\pv$. We call $a_i\pv$ the \textit{order component} of $\E_i$ and $\r_i$ its \textit{readout component}, and likewise for $\U_i$. Section~\ref{sec:theory} establishes conditions for this dissociation, which we evaluate empirically in Secs.~\ref{sec:geometry} and \ref{sec:causal}. In experiments, $\pv$ is the first principal component (PC1) of $\E$, oriented to have positive association with $i$; the readout component lies in its orthogonal complement.

\section{Geometric Conditions for the Max Task}
\label{sec:theory}

We first derive geometric constraints imposed by comparison and identity readout. In the saturated regime, where attention concentrates on the tokens with the largest score \citep{merrill2021effectsparameternormgrowth, hu2026transformerapproximationsrelus}, attention retrieves the embedding of the maximum and the readout decodes it, which requires
\begin{align}
\label{eq:max-condition}
    \text{(max-selection)} \qquad \E_q\cdot\E_m &> \E_q\cdot\E_k, \qquad k<m,\;q\leq m, \\
\label{eq:readout-condition}
    \text{(identity readout)} \qquad \E_m\cdot\U_m &> \E_m\cdot\U_j, \qquad j\neq m.
\end{align}
The maximum must receive the highest attention score, and its retrieved embedding must assign the highest logit to the corresponding output class.

\begin{theorem}[Geometric Implications]\label{thm:geometric-conditions}
For nonzero vectors, these conditions imply, for any two numbers $i<j$,
\begin{align}
\label{eq:norm-increase}
\text{(norm increase)} \qquad &\|\E_i\|_2 < \|\E_j\|_2, \\
\label{eq:self-similarity}
\text{(self-similarity)} \qquad &\cos\theta(\E_i,\E_j) > \frac{\|\E_i\|_2}{\|\E_j\|_2}, \\
\label{eq:cross-alignment}
\text{(cross-alignment)} \qquad &\cos\theta(\E_i,\U_j) < \cos\theta(\E_i,\U_i)\frac{\|\U_i\|_2}{\|\U_j\|_2}.
\end{align}
If adjacent numbers satisfy $\E_i\cdot\E_{i+1}\geq\kappa_i\|\E_i\|_2^2$ with $\kappa_i\geq\kappa>1$, then $\|\E_j\|_2\geq\kappa^{j-i}\|\E_i\|_2$.
\end{theorem}

\paragraph{Intuition.} Because the query is also a key, a smaller query must assign a higher score to a larger number than to itself. This requires increasing embedding norms and imposes a lower bound on angular similarity (Eqs.~\eqref{eq:norm-increase}--\eqref{eq:self-similarity}). Identity readout additionally constrains embedding--unembedding alignment (Eq.~\eqref{eq:cross-alignment}).

\begin{figure}[t]
\centering
    \includegraphics[width=0.94\linewidth]{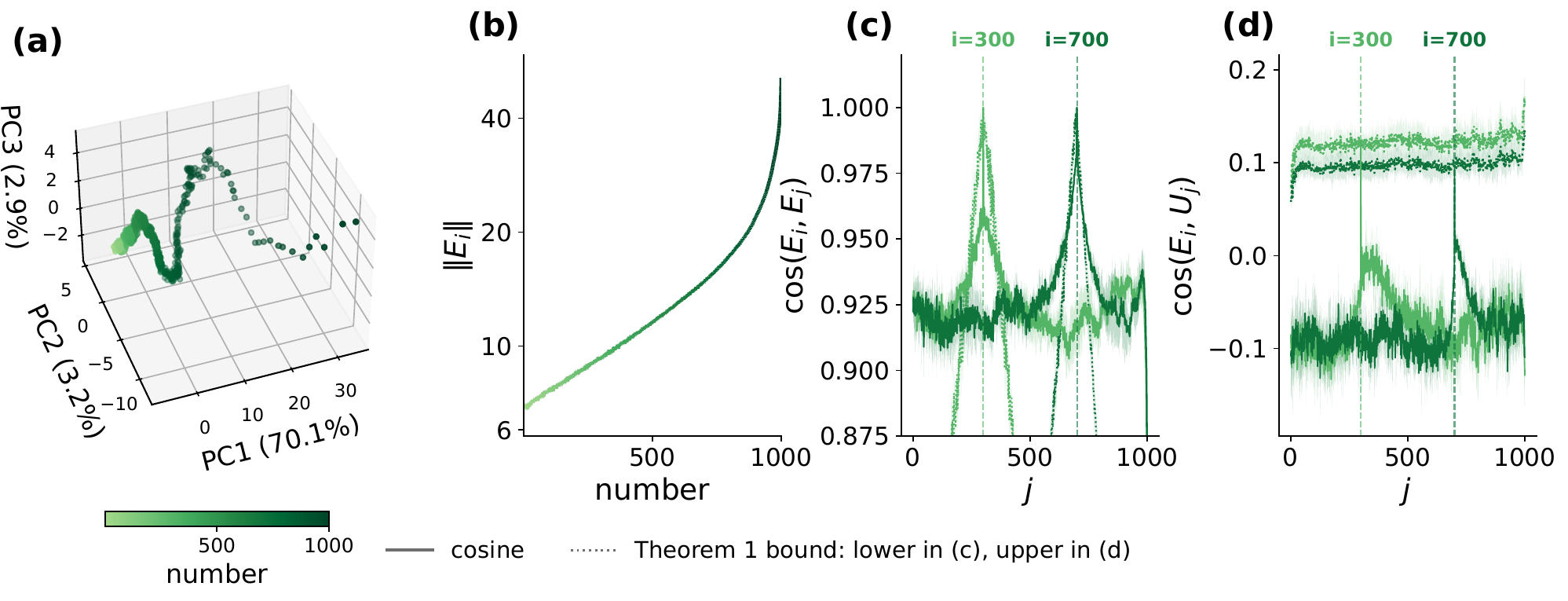}
    \caption{\textbf{Learned embeddings largely satisfy the conditions of Theorem~\ref{thm:geometric-conditions}.} \textbf{(a)} The top three PCs of $\E$ reveal a curved manifold elongated along PC1. \textbf{(b)} Embedding norm grows with number, as Eq.~\eqref{eq:norm-increase} requires. \textbf{(c)} $\cos(\E_i,\E_j)$ for $i=300$ and $700$ largely exceeds the symmetric lower bound associated with Eq.~\eqref{eq:self-similarity} (dotted; App.~\ref{app:running-max-metrics}). \textbf{(d)} $\cos(\E_i,\U_j)$ peaks at $j=i$ and remains below the upper bound for $j\neq i$ of Eq.~\eqref{eq:cross-alignment} (dotted). Mean and SD over four seeds; (a): one seed.}
    \label{fig:geometry-conditions}
\end{figure}

\begin{theorem}[Norm-efficient max selection]\label{thm:norm-efficient-selection}
Fix $\E_1\neq0$ and adjacent multiplicative margins $\kappa_i>1$. Among embeddings satisfying $\E_i\cdot\E_{i+1}\geq\kappa_i\|\E_i\|_2^2$, the solution minimizing $\sum_{i=2}^V\|\E_i\|_2^2$ is a positive rank-one ordered ray $\E_i=a_i\pv$, with $a_i$ increasing in $i$. However,
\begin{equation}
    \arg\max_j \E_i\cdot\U_j=\arg\max_j \pv\cdot\U_j,
\end{equation}
independent of $i$. Thus the bias-free linear readout used here cannot decode multiple identities from the rank-one solution.
\end{theorem}

\paragraph{Intuition.} The norm-minimizing representation encodes order in the norms of embeddings along a common ray. Since a bias-free linear decoder predicts the same class for all embeddings on this ray, identity readout requires an additional component. This requirement does not imply curvature: an affine line not passing through the origin also suffices (App.~\ref{sec:theory-appendix}). Moreover, similar curved geometry emerges with a class-biased decoder, for which the impossibility result does not apply (App.~\ref{app:ablations}). Its form may instead reflect properties of the data (Prop.~\ref{prop:translation-symmetry}).

\begin{theorem}[Functional dissociation of order and identity]\label{thm:functional-dissociation}
For a query $q$, the maximum $m$ and another key $k$,
\begin{equation}
\label{eq:selection-split}
    \E_q\cdot\E_m-\E_q\cdot\E_k
    =a_q(a_m-a_k)+\r_q\cdot(\r_m-\r_k).
\end{equation}
Thus the order component determines attention ordering if
\begin{equation}
\label{eq:budget}
    a_q(a_m-a_k)>|\r_q\cdot(\r_m-\r_k)|.
\end{equation}
After selecting $m$,
\begin{equation}
\label{eq:readout-split}
    \E_m\cdot(\U_m-\U_j)=a_m(b_m-b_j)+\r_m\cdot(\s_m-\s_j),
\end{equation}
so the readout component determines identity if $\r_m\cdot(\s_m-\s_j)>|a_m(b_m-b_j)|$ for $j\neq m$.
\end{theorem}

\paragraph{Intuition.} The attention margin decomposes into an \textit{order term} and an \textit{interference term} arising from the readout components. When the order term dominates (the \textit{interference budget}, Eq.~\eqref{eq:budget}), the readout component cannot reverse the comparison outcome. Together with the readout dominance condition, this establishes the feasibility of a clay-like functional dissociation. These conditions are sufficient, not necessary, and do not imply that training will recover such a decomposition.

\section{The Learned Geometry Separates Order and Identity}
\label{sec:geometry}

\paragraph{Empirical satisfaction of the geometric conditions.} We observe that embedding norms grow with number: Eqs.~\eqref{eq:norm-increase} and \eqref{eq:cross-alignment} hold for over 99\% of pairs of numbers, and the symmetric diagnostic for Eq.~\eqref{eq:self-similarity} holds for 92\% (Fig.~\ref{fig:geometry-conditions}b--d; App.~\ref{app:running-max-metrics}). Violations concentrate at smaller numbers, which occur infrequently as running maxima during training and receive less concentrated attention (App.~Fig.~\ref{fig:app-theorem1}).

\begin{figure}[t]
\centering
    \includegraphics[width=0.66\linewidth]{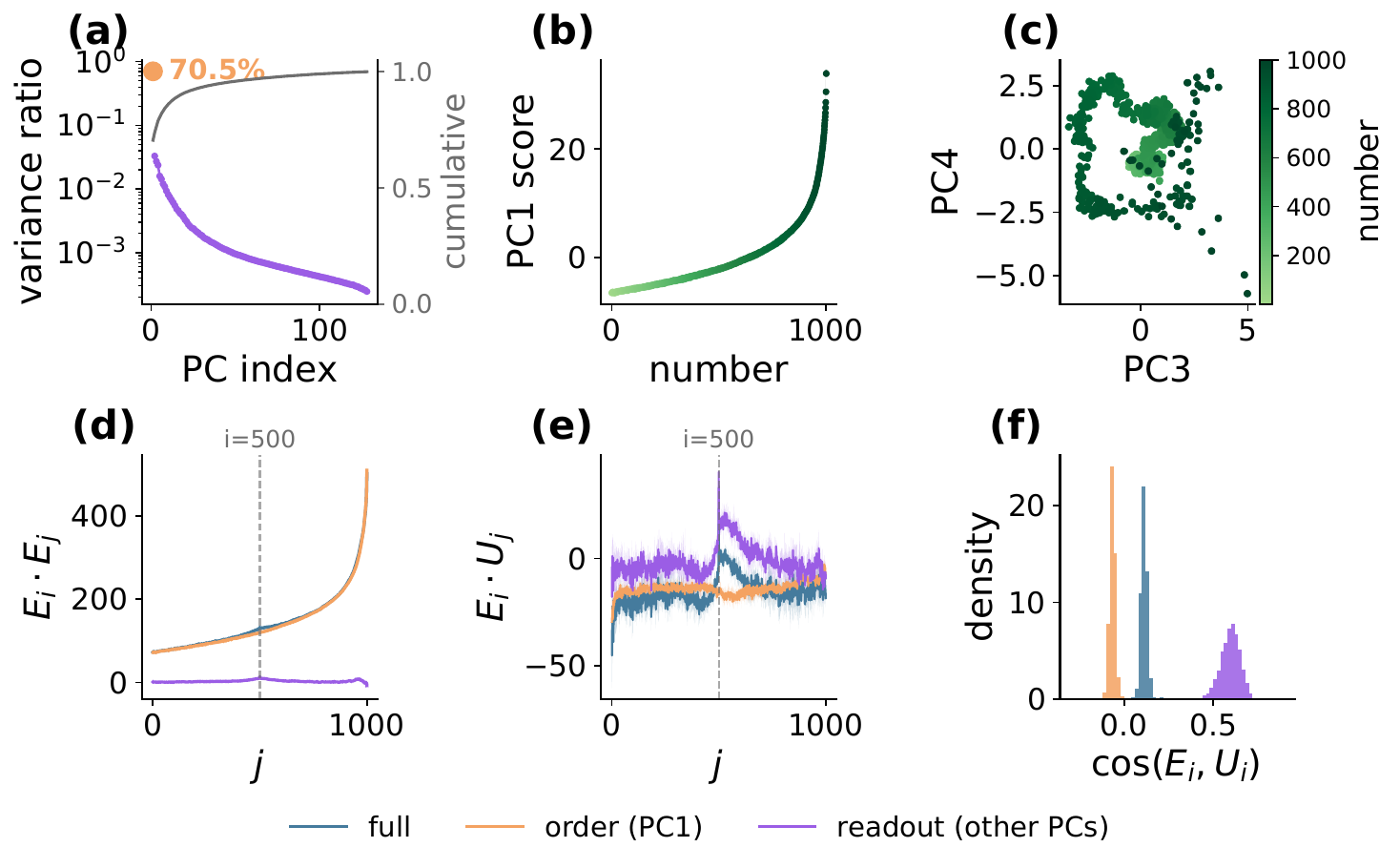}
    \caption{\textbf{PC1 encodes order, while the readout component supports embedding--unembedding alignment.} \textbf{(a, b)} PC1 explains 70\% of the variance; its scores increase with number. \textbf{(c)} The readout component exhibits curved geometry (PC3 versus PC4). \textbf{(d, e)} Decomposition of inner products ($i=500$): the order component dominates $\E_i\cdot\E_j$, while the readout component accounts for the peak of $\E_i\cdot\U_j$ at $j=i$. \textbf{(f)} Alignment between $\E_i$ and $\U_i$ is concentrated in the readout component. Mean and SD over four seeds; (c, f): one seed.}
    \label{fig:info-identity-separation}
\end{figure}

\paragraph{Identification of the order component.} PC1 explains 70\% of the variance of $\E$, and its scores increase almost monotonically with number (Fig.~\ref{fig:info-identity-separation}a,b). The order component accounts for nearly all the variation in comparison scores $\E_i\cdot\E_j$ (Fig.~\ref{fig:info-identity-separation}d), and PC1 closely aligns with the mean-difference direction estimated independently of PCA (App.~\ref{app:running-max-metrics}).

\paragraph{Identity readout.} The readout component exhibits structured, nonlinear geometry (Fig.~\ref{fig:info-identity-separation}c) and contributes relatively little to $\E_i\cdot\E_j$, but accounts for the peak of $\E_i\cdot\U_j$ at $j=i$ required for correct decoding (Fig.~\ref{fig:info-identity-separation}e,f). The decomposition emerges early in training (App.~Fig.~\ref{fig:app-epochs}).

\begin{figure}[t]
\centering
    \includegraphics[width=0.7\linewidth]{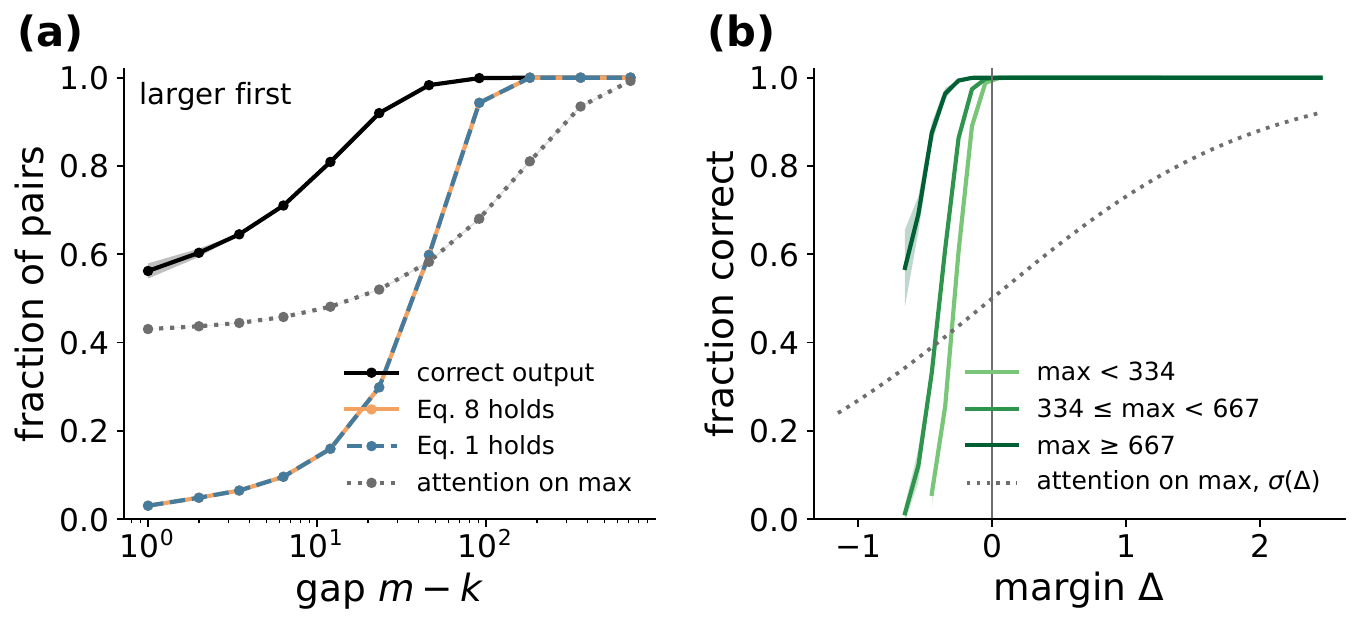}
    \caption{\textbf{On two-token inputs, the comparison margin predicts the model's errors.} \textbf{(a)} Larger number first: Eqs.~\eqref{eq:max-condition} and \eqref{eq:budget} hold for nearly identical fractions of pairs and frequently fail at small gaps, where attention on the maximum falls below one half. Accuracy nevertheless exceeds the fraction satisfying either condition. \textbf{(b)} Accuracy as a function of the margin $\Delta=(\E_k\cdot\E_m-\E_k\cdot\E_k)/\sqrt d$, stratified by the maximum: correct predictions occur even where attention on the maximum, $\sigma(\Delta)$ (dotted), is below one half. Mean and SD over four seeds.}
    \label{fig:pairs-main}
\end{figure}

\paragraph{Predicting comparison errors from geometry.}
\label{sec:bridge}
Attention is more concentrated for large maxima than for small maxima separated from competitors by narrow gaps (App.~Fig.~\ref{fig:app-saturation}). Two-token inputs permit an exact characterization of attention on the maximum as $\sigma(\Delta)$, where $\sigma$ is the logistic function and $\Delta$ is the score difference in Eq.~\eqref{eq:selection-split} divided by $\sqrt d$ (App.~\ref{app:pairs}). Errors occur predominantly when the larger number precedes the smaller one, requiring the query to assign a higher score to the earlier maximum than to itself (Fig.~\ref{fig:pairs-main}a). The order term alone predicts which of these pairs the model answers correctly, with an area under the ROC curve of 0.993, compared with 0.979 for the numerical gap. In this regime, the readout component interferes with comparison: its contribution to the margin is negative in over 99\% of pairs, increasing attention to the query relative to the maximum. Nevertheless, the model often predicts correctly at negative margins, when attention on the maximum is below one half (Fig.~\ref{fig:pairs-main}b), indicating that the decoder can recover identity from mixtures of embeddings outside the saturated-attention regime.

\section{The Two Components Have Distinct Causal Effects}
\label{sec:causal}

\begin{figure}[t]
    \centering
    \includegraphics[width=0.7\linewidth]{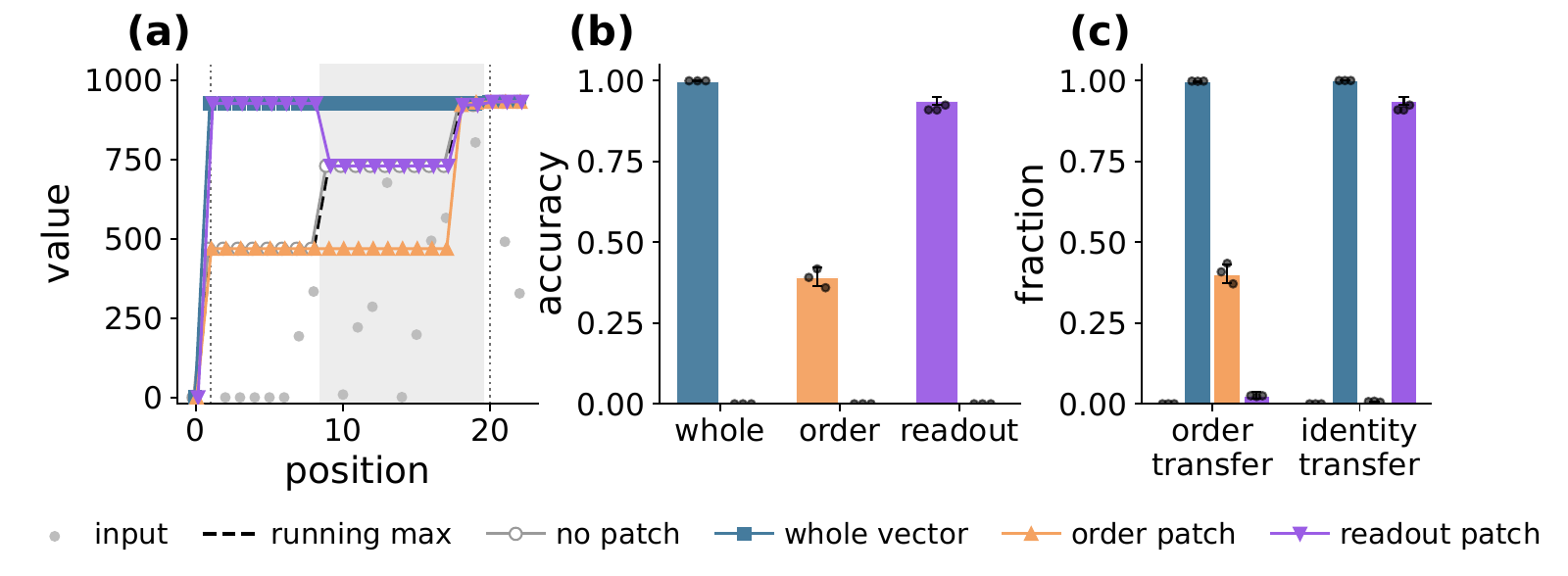}
    \caption{\textbf{Selective causal effects of the order and readout patches.} \textbf{(a)} Example intervention on the second token (zero-based index 1) using a larger source value $v_{\mathrm{new}}$ (shaded: a later token exceeds $v_{\mathrm{orig}}$). The order patch prolongs selection of $v_{\mathrm{orig}}$, whereas the readout patch substitutes $v_{\mathrm{new}}$ without altering the selection interval. \textbf{(b)} Accuracy of each patch against its own target, where that target differs from the unpatched output (light bars: no patch). \textbf{(c)} Order and identity transfer for each patch. Bars: one seed with 95\% bootstrap CIs; dots: three other seeds.}
    \label{fig:steering-results}
\end{figure}

\paragraph{Setup.} To assess the causal roles of these components, we perform three interchange interventions on a token embedding \citep{geiger2021causal, geiger2024finding}: the whole vector, $\bm h'=\hsrc$; the order component only (\textit{order patch}), $\bm h'=(I-P)\bm h+P\hsrc$; and the readout component only (\textit{readout patch}), $\bm h'=(I-P)\hsrc+P\bm h$, with $P=\pv\pv^\top$. In held-out sequences, the target is the second token (index 1 in the zero-based convention of this experiment), with value $v_{\mathrm{orig}}$, and the source has a larger value $v_{\mathrm{new}}$ (Fig.~\ref{fig:steering-results}a; App.~\ref{app:running-max-novel-steer}). Under the functional dissociation of Sec.~\ref{sec:theory}, the order patch should alter selection of the patched token without changing its decoded identity, whereas the readout patch should alter identity without changing selection. We quantify \textit{order transfer} as the fraction of positions predicting $v_{\mathrm{orig}}$ or $v_{\mathrm{new}}$ after a later token exceeds $v_{\mathrm{orig}}$ but not $v_{\mathrm{new}}$, and \textit{identity transfer} as the fraction predicting $v_{\mathrm{new}}$ while $v_{\mathrm{orig}}$ remains the unpatched maximum.

\paragraph{Causal dissociation of order and identity.} The partial patches exhibit selective effects (Fig.~\ref{fig:steering-results}c): the order patch transfers order at 40\% of positions and identity at under 1\%, whereas the readout patch transfers identity at 92\% of positions and order at under 3\%; the whole vector transfers both almost perfectly. Patches along PC2 or random directions transfer neither, although they are smaller than patches along PC1, and a complementary intervention using a decoy number corroborates the dissociation (App.~\ref{app:more-results}). Incomplete order transfer is particularly pronounced at small gaps, consistent with non-saturated attention (App.~Fig.~\ref{fig:app-steering-controls}c).

\paragraph{Decodability versus causal relevance.} A ridge probe accurately decodes numerical value ($R^2=0.981$), despite its low alignment with the order direction (cosine similarity $0.216$; App.~\ref{app:running-max-metrics}). Yet interventions along the latter selectively affect comparison. Decodability alone therefore does not establish a direction's causal relevance to a particular operation.

These interventions provide causal evidence for the clay-like perspective: distinct components of the manifold can be manipulated to selectively modify different computational operations, without preserving the geometry of the manifold as a whole.

\section{Discussion and Limitations}
\label{sec:discussion}

\paragraph{Crystal-like and clay-like perspectives on manifold geometry.}
The causal efficacy of linear steering and manifold steering need not reflect competing accounts of representation geometry. Rather, their efficacy may depend on the computational operation under consideration. While manifold-preserving interventions may be appropriate for computations involving geometric transformations of manifolds \citep{gurnee2026modelsmanipulatemanifoldsgeometry}, our findings demonstrate that component-specific interventions can selectively modify individual operations without preserving the geometry of the manifold as a whole. Thus, the geometric structure that an intervention must preserve depends on the computation it seeks to modify.

\paragraph{From representational geometry to computational function.}
Although data statistics and packing constraints may explain the emergence of manifold geometry, its computational utility depends on the operations acting upon it. In our setting, the curved readout geometry is consistent with approximate translation symmetry (Prop.~\ref{prop:translation-symmetry}; App.~Fig.~\ref{fig:app-toeplitz}) and persists even when the decoder permits identity readout from a ray. Conversely, replacing dot-product attention with distance-based attention removes the dominant order direction (App.~\ref{app:ablations}). These observations distinguish the factors governing the emergence of representational geometry from those governing its computational role. Neither the presence of a geometric structure nor the information decodable from it establishes its computational role. We argue that geometric interpretability should therefore characterize representations in relation to the computational operations that act upon them, rather than characterize their geometry in isolation.

\paragraph{Limitations.}
Our results establish the feasibility and empirical emergence of a clay-like decomposition in a minimal model, rather than its prevalence in deep networks. The decoding impossibility in Theorem~\ref{thm:norm-efficient-selection} relies on a bias-free decoder, and our theory establishes sufficient conditions for functional dissociation in the saturated-attention regime. Empirically, this dissociation is imperfect, particularly for nearby numbers. Moreover, the origin of the approximate translation symmetry remains unexplained. These limitations motivate a broader question: do distinct computational requirements systematically induce functionally distinct geometric components within concept representations?

\acks{This work has been made possible in part by a gift from the Chan Zuckerberg Initiative Foundation to establish the Kempner Institute at Harvard University. SSRH thanks the Kempner Institute for access to compute resources. SSRH thanks Val\'erie Costa for insightful discussions and suggestions on improving the manuscript, and the CRISP group at Harvard for useful discussions and feedback.}

\bibliography{ref_neureps}

\newpage
\appendix

\section{Experimental Setup}
\label{app:expt-setup}

\subsection{Task and Data}
\label{app:running-max-task}

Each input sequence $x=(x_1,\dots,x_T)$ has tokens drawn i.i.d.\ uniformly from the $V$ numbers $\{1,\dots,V\}$, with no output-frequency balancing. The target at every position is
\begin{equation}
y_t = \max(x_1, \dots, x_t), \qquad t = 1, \dots, T.
\end{equation}
Because the tokens are i.i.d.\ uniform, the distribution of $y_t$ concentrates toward $V$ as $t$ grows, so large running maxima are much more common than small ones in natural sequences. This is one reason we also test all two-token inputs (App.~\ref{app:pairs}).

We use $V=1000$ and $T=100$, with two dataset sizes (fixed random seed for token sampling and for the train/validation split): $500{,}000$ sequences ($350{,}000$ train, $150{,}000$ validation) for the baseline and the distance-attention variant, and $50{,}000$ sequences ($35{,}000$ train, $15{,}000$ validation) for the distance-decoder variant. All accuracies and geometric statistics use the validation split that matches each model's training run.

\subsection{Model, Training and Seeds}
\label{app:running-max-model}

The model is the one in Sec.~\ref{sec:model} and Fig.~\ref{fig:model}: a learned embedding table, one causal self-attention block and a position-wise decoding head, with no positional encoding, residual connection or MLP. The query, key and value projections are fixed at the identity in every run, and the baseline decoder is linear without a bias term. Both $\E$ and $\U$ are full-rank, freely trained $V\times d$ tables with $d=128$. The loss is the token-wise cross-entropy between the logits and the running-max target, averaged over positions and the batch. The two variants of App.~\ref{app:ablations} replace either the attention scores or the decoder with a distance-based version.

All models are trained with Adam (learning rate $10^{-3}$, no weight decay, no gradient clipping, no learning-rate schedule) and batch size $32$, for 20 epochs (50 for the distance decoder, on the smaller dataset). All analyses use the final-epoch checkpoint, except the training-dynamics analysis. We train the baseline four times: the original run and three runs whose initial weights and data order are set by \texttt{torch.manual\_seed} with seeds 1, 2 and 3, with the same data and hyperparameters. We report the mean and standard deviation over the four runs, and use the original run where a figure shows a single model.

\subsection{Evaluation and Geometric Analysis}
\label{app:running-max-metrics}

\paragraph{Accuracy and uncertainty.} Accuracy is the fraction of correct top-1 running-maximum predictions over all (sequence, position) pairs of the full validation split. Confidence intervals are 95\% percentile bootstrap intervals with $10{,}000$ resamples of sequences (or examples), computed on the pooled ratio of correct to scored positions; differences between conditions use paired resamples \citep{efron1994bootstrap}.

\paragraph{PCA and the conditions of Theorem~\ref{thm:geometric-conditions}.} We fit PCA on the final-checkpoint table $\E$ with a full singular value decomposition, treating each number's embedding as one sample, and take $\pv$ to be the top component, oriented so that its score increases with number. Then $a_i=\E_i\cdot\pv$ and $\r_i=\E_i-a_i\pv$, as in Sec.~\ref{sec:model}. For Theorem~\ref{thm:geometric-conditions}, we count, for each anchor $i$, the fraction of numbers $j$ that satisfy each condition: Eq.~\eqref{eq:norm-increase} over $j>i$, and Eqs.~\eqref{eq:self-similarity} and \eqref{eq:cross-alignment} over $j\neq i$. For Eq.~\eqref{eq:self-similarity}, we use the symmetric diagnostic $\cos\theta(\E_i,\E_j)>\min(\|\E_i\|_2,\|\E_j\|_2)/\max(\|\E_i\|_2,\|\E_j\|_2)$. For $i<j$, this agrees with the theorem's bound whenever $\|\E_i\|_2\leq\|\E_j\|_2$; otherwise it is weaker. Pooled over pairs, Eq.~\eqref{eq:norm-increase}, the symmetric self-similarity diagnostic, and Eq.~\eqref{eq:cross-alignment} hold for approximately 99.3\%, 92\%, and 100\%, respectively, in every seed. The satisfaction rate of the original self-similarity inequality can therefore be at most about 0.7 percentage points lower than that of the symmetric diagnostic.

\paragraph{Fraction of variance explained (FVE).} For each anchor $i$, we measure how much of the curve $\E_i\cdot\E_j$ (over all $j$) is captured by the order component (PC1) versus the readout component,
\begin{equation}
\text{FVE}_c = 1 - \frac{SS_{\text{res},c}}{SS_{\text{tot}}}, \qquad
SS_{\text{tot}} = \sum_j \big(\E_i\cdot \E_j - \overline{\E_i\cdot \E_j}\big)^2,
\end{equation}
where $SS_{\text{res},c}$ for component $c$ is the sum of squared values of the \emph{other} component's dot-product curve. This is exact, since the two parts sum to the full curve at every $j$. We report the median over all $V$ anchors.

\paragraph{Order directions found without PCA.} We compare PC1 with two directions that do not use PCA: the difference between the mean embeddings of the larger and the smaller half of the numbers, and the coefficient vector of a ridge regression from $\E_i$ to $i$ (regularization chosen by generalized cross-validation; $R^2$ from 5-fold cross-validation). In the original run, the ridge probe achieves $R^2=0.981$ with cosine similarity $0.216$ between its coefficient vector and PC1. For Prop.~\ref{prop:pc1-recovery}, we use the mean-difference direction as $\pv$ and report the ratio of Eq.~\eqref{eq:app-pc1-condition} and the normalized correlation $\|\c\|_2/(\operatorname{sd}(\widetilde a)\sqrt{\lambda_{\max}})$, which Eq.~\eqref{eq:app-uncorrelated} assumes is zero.

\subsection{Causal Interventions}
\label{app:running-max-steering}

Both experiments intervene on a token embedding, the only intervention site in this architecture, with the three patches of Sec.~\ref{sec:causal} and $P=\pv\pv^\top$. These are rank-restricted interchange interventions with a PCA-defined subspace, in contrast to Distributed Alignment Search \citep{geiger2024finding}, which learns the intervention subspace.

\subsubsection{Interchange intervention with a decoy}
\label{app:running-max-iia}

We scan the validation set for natural running-max transitions: a value $v$ becomes the running max at position $t_v\geq 5$, holds it for at least $3$ steps, and is then overtaken, and a previous maximum $v_{\text{prev}}$ exists. The \emph{clean} input is the unmodified sequence. The \emph{corrupted} input replaces the token at $t_v$ with a decoy $v_c=\max(1,v_{\text{prev}}-10)$, so its running max stays at $v_{\text{prev}}$. We patch the embedding at $t_v$ of the corrupted forward pass with source $h_{\mathrm{src}}=\E_v$ and read the logits at $t_v+2$, before the real transition, so a correct read-out requires the patched information to pass through attention. We report the interchange intervention accuracy (IIA; \citealt{geiger2021causal}), the fraction of examples whose top-1 prediction equals $v$, and more generally the fractions whose prediction equals $v$, $v_c$ or $v_{\text{prev}}$. The whole-vector patch should output $v$, and the order patch should output $v_c$, because the patched token then takes the source's order and keeps the decoy's identity. We also report the median logit-margin recovery, $(ld_p-ld_{\text{corr}})/(ld_{\text{clean}}-ld_{\text{corr}})$, where $ld$ is the logit difference between $v$ and $v_{\text{prev}}$ in the patched, corrupted and clean runs (0: no effect; 1: full recovery), excluding one example with $|ld_{\text{clean}}-ld_{\text{corr}}|<10^{-3}$. We use $1000$ held-out examples per patch.

\subsubsection{Direct steering to a novel value}
\label{app:running-max-novel-steer}

For this experiment only, we use zero-based token indices (as in the intervention plots). Each example is constructed from a validation sequence $x$, with the second token (index $1$) as the patch site and window length $T_{\text{over}}=20$: (1) the first token ($x_0$) is set to the smallest number, so $x_1$ is a running maximum; (2) a third of the positions after position $1$ are forced below $x_1$'s original value $v_{\text{orig}}$, so it survives as the running max for a stretch; (3) $v_{\text{new}}=\max(x_{1:T_{\text{over}}})+5$, larger than every value in the window, is the value steered into position $1$; and (4) $x_{T_{\text{over}}}\leftarrow v_{\text{new}}+5$, so $v_{\text{new}}$ is overtaken on schedule. Sequences for which $v_{\text{new}}+5$ would exceed the largest number are skipped, and positions beyond $T_{\text{over}}$ are untouched. We construct $500$ held-out examples and one demonstration example, and patch position $1$ with source $h_{\mathrm{src}}=\E_{v_{\text{new}}}$.

Let $y^{\text{patched}}$ be the running max if the input had contained $v_{\text{new}}$ at position $1$, and $y^{\text{true}}$ the unpatched running max. The \emph{whole-vector} target is $y^{\text{patched}}$. The \emph{order-patch} target is $y^{\text{patched}}$ with $v_{\text{new}}$ relabeled $v_{\text{orig}}$: the patch should change \emph{when} the maximum changes but not the value read out. The \emph{readout-patch} target is $y^{\text{true}}$ with $v_{\text{orig}}$ relabeled $v_{\text{new}}$: the trajectory should stay the same, but the value read out should switch to $v_{\text{new}}$. Accuracy is computed over $t=1,\dots,T_{\text{over}}$, the only positions that can tell the three hypotheses apart.

\paragraph{Changed positions.} Many positions in the window have the same target with and without a patch, where the unpatched model already scores well. We therefore report each patch's accuracy over the positions where its target differs from the unpatched running max. There the no-patch accuracy is 0, so effects are directly comparable across patches (Fig.~\ref{fig:app-steering-controls}a).

\paragraph{Order and identity read-outs.} In stretch A, the unpatched running max is still $v_{\text{orig}}$; in stretch B, a natural token has overtaken $v_{\text{orig}}$ before $T_{\text{over}}$, but not $v_{\text{new}}$. \emph{Order taken from the source} is the fraction of stretch-B positions whose output still comes from the patched position ($v_{\text{new}}$ or $v_{\text{orig}}$). \emph{Identity taken from the source} is the fraction of stretch-A positions whose output is $v_{\text{new}}$.

\paragraph{Controls and failure analysis.} We repeat the two partial patches with PC1 replaced by PC2 or by each of 100 random unit directions (fixed random seed), and record both read-outs and the mean patch size along the direction. Because numbers differ mostly along PC1, patches along PC2 and random directions are smaller on average (mean size 2.2 and 1.1, against 12.5 along PC1). For each stretch-B position, we also check whether the hybrid vector $(I-P)\E_{v_{\text{orig}}}+P\E_{v_{\text{new}}}$ receives the highest attention score, and record the gap between $v_{\text{new}}$ and the competing natural maximum; confidence intervals for this analysis resample sequences.

\paragraph{A note on the construction.} The construction sets up to seven tokens to the smallest number, which is the weakest query. Without any patch, the model outputs the correct value at 71\% of stretch-A positions whose query is the smallest number, against 98\% elsewhere. The changed-position metric avoids this effect, because the no-patch baseline is 0 there.

\subsection{Saturation, Interference, Symmetry and Training Dynamics}
\label{app:other-methods}

\paragraph{Attention saturation.} On the first 2000 validation sequences, for every position whose prefix contains a token smaller than the running max $m$, we record the attention mass on all copies of $m$, the gap between $m$ and the next-largest earlier value, $m$ itself, whether the output is correct, whether Eq.~\eqref{eq:max-condition} holds (the maximum receives the strictly highest score), and whether Eq.~\eqref{eq:budget} holds against every smaller earlier token.

\paragraph{Conditions over the vocabulary.} We evaluate Eqs.~\eqref{eq:max-condition} and \eqref{eq:budget} over all vocabulary triples $(q,m,k)$ with $k<m$ and $q\leq m$, about $3.3\times10^{8}$ triples. On the readout side, we compute for each $m$ the fraction of $j\neq m$ for which Eq.~\eqref{eq:readout-condition} holds and for which the readout component alone decides it, Eq.~\eqref{eq:app-readout-dominance}.

\paragraph{Translation symmetry.} For Prop.~\ref{prop:translation-symmetry}, we center the readout components $\r_i$ and fit both their Gram matrix and their cosine-similarity matrix with a Toeplitz model $G_{ij}\approx g(|i-j|)$, where $g$ is the mean of each diagonal. We report $R^2$ over off-diagonal entries, for all numbers and for numbers below 800, and compare it with the largest $R^2$ over 20 shuffles of the number labels.

\paragraph{Training dynamics.} For every saved epoch of the original run, we compute accuracy on 2000 validation sequences, the PC1 variance ratio, $|\cos|$ between PC1 at that epoch and at the final epoch, and the order and identity read-outs of the three patches.

\section{Analytical Derivations}
\label{app:derive-theory}

Throughout this appendix, $i,j$ denote generic numbers; within one comparison, $q$ is the query, $m$ the running maximum and $k$ any other key, so that $k<m$ and $q\leq m$.

\subsection{Why numbers?}
\label{app:why-numbers}

Consider a finite set $\setS$, with $|\setS|=V$, equipped with a total order $\leq$. For every non-empty subset $\C\subseteq\setS$, $\max(\C)$ is the unique $y\in\C$ such that $x\leq y$ for all $x\in\C$. Learning the maximum of arbitrary non-empty subsets of $\setS$ amounts to learning the total order of $\setS$: all two-element subsets occur, and their maxima recover every pairwise comparison. Many object sets become totally ordered after choosing an attribute, such as weight, size, temperature, age or hue. Any finite totally ordered set admits an order-preserving bijection to $\{1,\dots,V\}$, so we use natural numbers, and the running-maximum task is a minimal comparison problem.

\subsection{Saturated max selection and Theorem~\ref{thm:geometric-conditions}}

Under identity query and key matrices, the score that the query $q$ gives an earlier token $k$ is proportional to $\E_q\cdot\E_k$. In the saturated limit, attention is supported only on the tokens with the largest score, so retrieving the maximum $m$ requires Eq.~\eqref{eq:max-condition} for every smaller key in the sequence; the same condition separates repeated copies of the maximum from all other positions. After selection, the attention output is $\E_m$, the logit for output $j$ is $\E_m\cdot\U_j$, and correct decoding requires Eq.~\eqref{eq:readout-condition}. Since every number is the running maximum of some sequence, Eq.~\eqref{eq:readout-condition} applies to every $m$.

\paragraph{Proof of Theorem~\ref{thm:geometric-conditions}.}
Consider a sequence in which $j$ is the running maximum and $i<j$ is the query. Since the query is itself an available attention target, selecting $j$ requires
\begin{equation}
    \E_i\cdot\E_j>\E_i\cdot\E_i=\|\E_i\|_2^2.
    \label{eq:app-self-competitor}
\end{equation}
By Cauchy--Schwarz, $\E_i\cdot\E_j\leq\|\E_i\|_2\|\E_j\|_2$, so $\|\E_j\|_2>\|\E_i\|_2$, which is Eq.~\eqref{eq:norm-increase}. Writing $\E_i\cdot\E_j=\|\E_i\|_2\|\E_j\|_2\cos\theta(\E_i,\E_j)$ gives Eq.~\eqref{eq:self-similarity}. For adjacent numbers, the margin constraint and Cauchy--Schwarz give $\|\E_{l+1}\|_2\geq\kappa_l\|\E_l\|_2$, and iterating from $i$ to $j$ gives
\begin{equation}
    \|\E_j\|_2 \geq \|\E_i\|_2\prod_{l=i}^{j-1}\kappa_l,
    \label{eq:app-norm-product}
\end{equation}
which reduces to the bound in Theorem~\ref{thm:geometric-conditions} when $\kappa_l\geq\kappa$. Finally, correct readout of $i$ requires $\E_i\cdot\U_i>\E_i\cdot\U_j$ for $j\neq i$. Writing both sides with norms and angles and cancelling the positive factor $\|\E_i\|_2$ gives Eq.~\eqref{eq:cross-alignment}.
\hfill$\square$

When the query is itself the maximum ($q=m$), the max-selection condition reads $\|\E_m\|_2^2>\E_m\cdot\E_k$ for $k<m$. Given Eq.~\eqref{eq:norm-increase}, this follows from Cauchy--Schwarz, so it adds no further constraint. The constraints of Theorem~\ref{thm:geometric-conditions} all come from queries that are smaller than the maximum.

\begin{proposition}[Finite-softmax selection margin]\label{prop:finite-softmax}
Suppose $q<m$ and the unique maximum $m$ receives attention probability at least $1-\epsilon$, with $\epsilon<1/2$. Then
\begin{equation}
    \E_q\cdot\E_m-\|\E_q\|_2^2
    \geq \sqrt d\log\frac{1-\epsilon}{\epsilon}.
    \label{eq:softmax-additive-margin}
\end{equation}
Equivalently, defining $\kappa_{q,m}=(\E_q\cdot\E_m)/\|\E_q\|_2^2$,
\begin{equation}
    \kappa_{q,m}\geq 1+\frac{\sqrt d}{\|\E_q\|_2^2}\log\frac{1-\epsilon}{\epsilon}.
\end{equation}
\end{proposition}

\begin{proof}
Let $\gamma_k=(\E_q\cdot\E_k)/\sqrt d$ denote the score of key $k$. If the softmax probability of $m$ is at least $1-\epsilon$, then
\begin{equation}
    \gamma_m-\log\sum_{k\neq m}e^{\gamma_k}\geq\log\frac{1-\epsilon}{\epsilon}.
\end{equation}
Because $q<m$, the query $q$ is a distinct competitor, hence $\log\sum_{k\neq m}e^{\gamma_k}\geq \gamma_q$, and therefore $\gamma_m-\gamma_q\geq\log\frac{1-\epsilon}{\epsilon}$. This gives Eq.~\ref{eq:softmax-additive-margin}, and dividing by $\|\E_q\|_2^2$ gives the bound on $\kappa_{q,m}$.
\end{proof}

Thus, for a fixed degree of saturation, the multiplicative margin above one can decrease as embedding norms increase.

\subsection{Norm efficiency and the need for a readout component}
\label{sec:theory-appendix}

\paragraph{Proof of Theorem~\ref{thm:norm-efficient-selection}.}
For every embedding table satisfying the adjacent margins, Cauchy--Schwarz gives $\|\E_{i+1}\|_2\geq\kappa_i\|\E_i\|_2$, so inductively $\|\E_i\|_2\geq\big(\prod_{l=1}^{i-1}\kappa_l\big)\|\E_1\|_2$ for all $i$. The objective $\sum_{i=2}^V\|\E_i\|_2^2$ therefore has a termwise lower bound, and the construction $\E_{i+1}=\kappa_i\E_i$ attains every bound at once, so the minimizer is
\begin{equation}
    \E_j=\left(\prod_{i=1}^{j-1}\kappa_i\right)\E_1.
    \label{eq:app-minnorm-ray}
\end{equation}
Equality in Cauchy--Schwarz with a positive inner product forces positive collinearity, so the minimizer is unique. Writing $\E_i=a_i\pv$ with $a_i>0$ increasing in $i$, every query $q$ satisfies $\E_q\cdot\E_m=a_qa_m>a_qa_k=\E_q\cdot\E_k$ for $m>k$, so the ray also satisfies every pairwise comparison. For any unembedding $\U$, the bias-free logit of class $j$ given the embedding of number $i$ is $\E_i\cdot\U_j=a_i\,\pv\cdot\U_j$. Since $a_i>0$, $\arg\max_j\E_i\cdot\U_j=\arg\max_j\pv\cdot\U_j$ for every $i$, so the readout outputs the same class for every input. With class-dependent biases, the affine logits can cross as $a_i$ varies, and this conclusion no longer holds.
\hfill$\square$

The theorem is an extremal statement about margin efficiency, not a claim that our unregularized training objective minimizes $\sum_i\|\E_i\|_2^2$.

\paragraph{An offset line also decodes identity.} The impossibility in Theorem~\ref{thm:norm-efficient-selection} applies only to a ray through the origin. Take increasing positive $a_i$ and the two-dimensional embeddings $\E_i=(a_i,1)$, which lie on a straight line offset from the origin, with unembeddings $\U_j=(2a_j,-a_j^2)$. Max selection holds because $\E_q\cdot\E_m-\E_q\cdot\E_k=a_q(a_m-a_k)>0$ whenever $m>k$, and the bias-free logits $\E_i\cdot\U_j=2a_ia_j-a_j^2=a_i^2-(a_i-a_j)^2$ are uniquely maximized at $j=i$. The constant second coordinate acts as a readout bias, so identity can be decoded without any curvature. Structure off the ray is needed for a bias-free readout, but the theory does not require it to be curved.

\paragraph{Corollary (embeddings and unembeddings cannot be tied).}
A saturated retrieve-then-decode solution that computes running maximum over the full vocabulary cannot satisfy $\U=\E$.

\begin{proof}
Take $q<m$. On a sequence where $m$ is the maximum and $q$ is the query, max-selection requires $\E_q\cdot\E_m>\E_q\cdot\E_q$. On a sequence where $q$ itself is the maximum, correct decoding requires $\E_q\cdot\U_q>\E_q\cdot\U_m$. If $\U=\E$, the second condition becomes $\E_q\cdot\E_q>\E_q\cdot\E_m$, which contradicts the first.
\end{proof}

\subsection{Functional decomposition of order and identity}
\label{app:decomposition}

For any unit vector $\pv$, write $\E_i=a_i\pv+\r_i$ and $\U_i=b_i\pv+\s_i$ with $\r_i,\s_i\perp\pv$, the order and readout components of Sec.~\ref{sec:model}. The orthogonality here is definitional; the theorem gives conditions under which the two components control different parts of the computation.

\paragraph{Theorem \ref{thm:functional-dissociation} (restated).}
For a query $q$, the maximum $m$ and another key $k$,
\begin{equation}
    \E_q\cdot\E_m-\E_q\cdot\E_k
    =a_q(a_m-a_k)+\r_q\cdot(\r_m-\r_k).
    \label{eq:app-selection-decomposition}
\end{equation}
If
\begin{equation}
    a_q(a_m-a_k)>|\r_q\cdot(\r_m-\r_k)|,
    \label{eq:app-selection-dominance}
\end{equation}
then the order component determines the attention ordering. After saturated selection of $m$,
\begin{equation}
    \E_m\cdot(\U_m-\U_j)
    =a_m(b_m-b_j)+\r_m\cdot(\s_m-\s_j).
    \label{eq:app-readout-decomposition}
\end{equation}
If
\begin{equation}
    \r_m\cdot(\s_m-\s_j)>|a_m(b_m-b_j)|,
    \qquad \forall j\neq m,
    \label{eq:app-readout-dominance}
\end{equation}
then the readout component determines the decoded identity.

\begin{proof}
Orthogonality gives $\E_q\cdot\E_m=a_qa_m+\r_q\cdot\r_m$ and the analogous expression for $k$, giving Eq.~\ref{eq:app-selection-decomposition}. Under Eq.~\ref{eq:app-selection-dominance}, the score difference is at least $a_q(a_m-a_k)-|\r_q\cdot(\r_m-\r_k)|>0$, so the order component alone determines which candidate wins attention. The readout margin expands in the same way into Eq.~\ref{eq:app-readout-decomposition}, and under Eq.~\ref{eq:app-readout-dominance} it is at least $\r_m\cdot(\s_m-\s_j)-|a_m(b_m-b_j)|>0$, so the readout component alone determines the decoded identity.
\end{proof}

Equation~\ref{eq:app-selection-dominance} gives the readout component an explicit interference budget: the readout components need not have small norm, but their contribution cannot reverse the comparison margin. If $\|\r_i\|_2\leq R$ for all $i$, the simpler condition $a_q(a_m-a_k)>2R^2$ is also sufficient, but in the trained model it holds for only 0.03\% of vocabulary triples, while Eq.~\ref{eq:app-selection-dominance} holds for 99.25\%.

\paragraph{Sign of the interference when the query competes with itself.}
\label{app:self-bias}
When the query is smaller than the maximum, the query itself is one of the keys ($k=q$), and the interference is
\[
    \r_q\cdot(\r_m-\r_q)=\|\r_q\|_2\big(\|\r_m\|_2\cos\theta(\r_q,\r_m)-\|\r_q\|_2\big).
\]
It is negative exactly when $\cos\theta(\r_q,\r_m)<\|\r_q\|_2/\|\r_m\|_2$, the self-similarity bound of Eq.~\eqref{eq:self-similarity} reversed and applied to the readout components. If the interference is negative, Eq.~\ref{eq:app-selection-dominance} reduces to $a_q(a_m-a_q)+\r_q\cdot(\r_m-\r_q)>0$, which is the max-selection condition itself, so Eq.~\ref{eq:app-selection-dominance} is then necessary as well as sufficient. When the query is the maximum ($q=m$), the interference against a smaller key $k$ is $\|\r_m\|_2^2-\r_m\cdot\r_k$, which is positive whenever $\cos\theta(\r_m,\r_k)<\|\r_m\|_2/\|\r_k\|_2$; the readout component then helps comparison.

\begin{proposition}[Recovery of the order direction by PCA]\label{prop:pc1-recovery}
Suppose the centered embeddings satisfy $\widetilde{\E}_i=\widetilde a_i\pv+\widetilde\r_i$ with $\widetilde\r_i\perp\pv$, and the order coordinate and the readout component are uncorrelated:
\begin{equation}
    \frac{1}{V}\sum_i\widetilde a_i\widetilde\r_i=0.
    \label{eq:app-uncorrelated}
\end{equation}
If
\begin{equation}
    \operatorname{Var}(\widetilde a)>
    \lambda_{\max}(\operatorname{Cov}(\widetilde\r)),
    \label{eq:app-pc1-condition}
\end{equation}
then $\pv$ is the unique first principal component direction of $\E$.
\end{proposition}

\begin{proof}
The embedding covariance is $\Sigma_{\E}=\operatorname{Var}(\widetilde a)\pv\pv^\top+\Sigma_{\r}+\pv\c^\top+\c\pv^\top$, where $\Sigma_{\r}=\operatorname{Cov}(\widetilde\r)$ and $\c=V^{-1}\sum_i\widetilde a_i\widetilde\r_i$. By Eq.~\ref{eq:app-uncorrelated}, $\c=0$, and since every $\widetilde\r_i\perp\pv$, $\Sigma_{\E}\pv=\operatorname{Var}(\widetilde a)\pv$. For any unit $\w\perp\pv$, $\w^\top\Sigma_{\E}\w=\w^\top\Sigma_{\r}\w\leq\lambda_{\max}(\Sigma_{\r})$. Under Eq.~\ref{eq:app-pc1-condition}, the eigenvalue along $\pv$ is strictly larger than every eigenvalue in $\pv^\perp$, so $\pv$ is the unique leading eigenvector.
\end{proof}

Thus the functional result does not itself require the order direction to be PC1; PCA recovers it when variation along the order direction dominates variation in the readout component and the two are uncorrelated. In the trained model, with $\pv$ taken as the mean-difference direction, the variance condition holds with a ratio of 16.7, but the normalized correlation is 0.62 rather than 0. A first-order perturbation argument then predicts a tilt of about $9^\circ$ between PC1 and $\pv$, which matches their measured cosine of 0.988.

\subsection{Translation symmetry and swirls}

The task conditions explain why a readout component is useful for identity, but they do not determine its geometry. Following the symmetry argument of \citet{karkada2026symmetry}, we can make a conditional statement. Let $\widetilde\r_i$ be the centered readout components and $G^{\perp}_{ij}=\widetilde\r_i\cdot\widetilde\r_j$ their Gram matrix.

\begin{proposition}[Translation symmetry implies Fourier modes]\label{prop:translation-symmetry}
Suppose the numbers are considered on a periodic one-dimensional lattice and the Gram matrix of the readout components is translation invariant,
\begin{equation}
    G^{\perp}_{ij}=g\!\left((i-j)\bmod V\right).
    \label{eq:residual-translation-symmetry}
\end{equation}
Then $G^{\perp}$ is circulant and its eigenvectors are discrete Fourier modes. Since $G^{\perp}$ is real and symmetric, its real eigenspaces can be represented by sine/cosine pairs. Consequently, the PC coordinates of the readout components are sinusoidal functions of number; matched sine/cosine PC pairs trace circles, while pairs of different frequencies trace Lissajous-like curves.
\end{proposition}

\begin{proof}
Equation~\ref{eq:residual-translation-symmetry} makes every row of $G^{\perp}$ a cyclic shift of the previous one, so $G^{\perp}$ is circulant, and circulant matrices are diagonalized by the discrete Fourier basis \citep{gray2006toeplitz}. Modes $f$ and $-f$ share an eigenvalue because $G^{\perp}$ is real and symmetric, so their real and imaginary parts, $\cos(2\pi fi/V)$ and $\sin(2\pi fi/V)$, span a real eigenspace. PCA coordinates come from the eigendecomposition of the centered Gram matrix, so the PCs of the readout components inherit these modes; a matched sine/cosine pair traces a circle as $i$ varies.
\end{proof}

Our vocabulary is an open number line rather than a cycle. The analogous condition $G^{\perp}_{ij}\approx g(|i-j|)$ gives an approximately Toeplitz matrix, whose eigenvectors are close to Fourier modes away from the boundaries \citep{gray2006toeplitz}. This is the symmetry-to-geometry mechanism of \citet{karkada2026symmetry}, applied conditionally to the readout component rather than derived from corpus statistics. We do not prove that the running-max data or task induces this symmetry. Empirically, the similarities between readout components of the trained model are close to Toeplitz for numbers below about 800 (Fig.~\ref{fig:app-toeplitz}).

\clearpage
\section{Additional Results for the Baseline Model}
\label{app:more-results}

Unless stated otherwise, bands and $\pm$ give the standard deviation over four seeds, and confidence intervals are 95\% bootstrap intervals for the original run.

\subsection{Robustness across seeds and the conditions of Theorem~\ref{thm:geometric-conditions}}

\begin{figure}[h]
    \centering
    \includegraphics[width=0.55\linewidth]{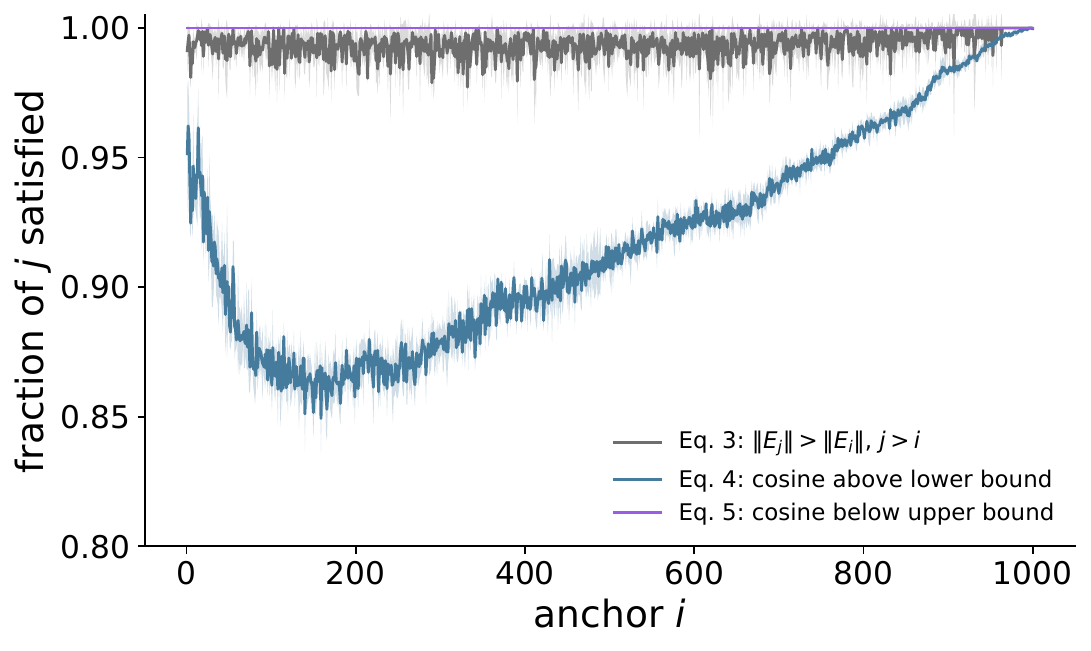}
    \caption{\textbf{The conditions of Theorem~\ref{thm:geometric-conditions} hold almost everywhere, except the self-similarity bound at small numbers.} For each anchor $i$, the fraction of numbers $j$ that satisfy Eqs.~\eqref{eq:norm-increase} and \eqref{eq:cross-alignment} and the symmetric diagnostic for Eq.~\eqref{eq:self-similarity} (App.~\ref{app:running-max-metrics}). The norm and cross-alignment conditions hold for nearly all eligible $j$. The self-similarity diagnostic fails for up to 14\% of $j$ at anchors between about 50 and 500, where embedding norms are small, as the finite-softmax bound (Prop.~\ref{prop:finite-softmax}) suggests. Mean and SD over four seeds.}
    \label{fig:app-theorem1}
\end{figure}

Table~\ref{tab:seeds} lists the main quantities for each training seed; all of them agree closely. Fig.~\ref{fig:app-theorem1} shows, anchor by anchor, where the conditions of Theorem~\ref{thm:geometric-conditions} hold.

\begin{table}[h]
\centering
\small
\begin{tabular}{lccccc}
\toprule
Quantity & Seed 0 & Seed 1 & Seed 2 & Seed 3 & Mean $\pm$ SD \\
\midrule
Accuracy (validation) & 0.9996 & 0.9996 & 0.9997 & 0.9997 & $0.9997\pm0.0000$ \\
PC1 variance ratio & 0.701 & 0.707 & 0.710 & 0.702 & $0.705\pm0.004$ \\
Spearman(PC1, number) & 0.9996 & 0.9996 & 0.9996 & 0.9996 & $0.9996\pm0.0000$ \\
FVE of $\E_i\cdot\E_j$ by PC1 (median) & 0.998 & 0.998 & 0.998 & 0.998 & $0.998\pm0.000$ \\
$|\cos|$(PC1, mean-difference direction) & 0.988 & 0.989 & 0.991 & 0.989 & $0.989\pm0.001$ \\
Order patch: order from source & 0.402 & 0.408 & 0.435 & 0.371 & $0.404\pm0.026$ \\
Order patch: identity from source & 0.003 & 0.007 & 0.008 & 0.005 & $0.006\pm0.002$ \\
Readout patch: order from source & 0.026 & 0.025 & 0.023 & 0.025 & $0.025\pm0.002$ \\
Readout patch: identity from source & 0.937 & 0.909 & 0.909 & 0.924 & $0.920\pm0.013$ \\
Attention on the max (median) & 0.976 & 0.979 & 0.980 & 0.979 & $0.979\pm0.002$ \\
Positions where Eq.~\eqref{eq:max-condition} holds & 0.993 & 0.994 & 0.993 & 0.993 & $0.993\pm0.001$ \\
Triples where Eq.~\eqref{eq:budget} holds & 0.992 & 0.993 & 0.993 & 0.992 & $0.993\pm0.000$ \\
Triples where Eq.~\eqref{eq:budget} holds, gap 1 & 0.449 & 0.452 & 0.448 & 0.441 & $0.448\pm0.005$ \\
Pairs where Eq.~\eqref{eq:app-readout-dominance} holds & 0.9996 & 0.9996 & 0.9994 & 0.9993 & $0.9995\pm0.0002$ \\
Toeplitz $R^2$ (cosine, numbers $<800$) & 0.816 & 0.801 & 0.786 & 0.820 & $0.806\pm0.015$ \\
\bottomrule
\end{tabular}
\caption{Main quantities for each of the four training seeds of the baseline model. Seed 0 is the original run.}
\label{tab:seeds}
\end{table}

\subsection{Controls for the causal interventions}

\begin{figure}[h]
    \centering
    \includegraphics[width=0.85\linewidth]{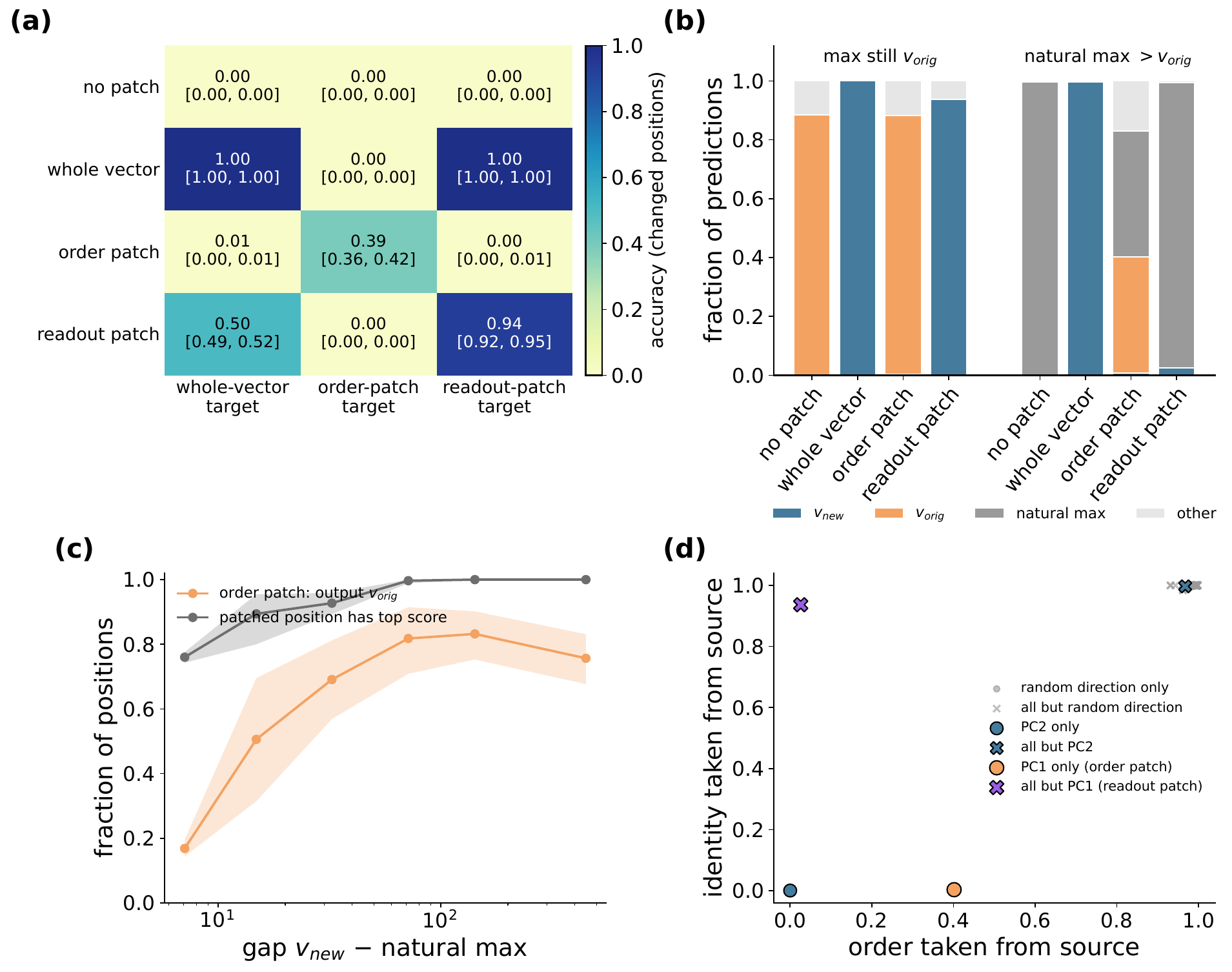}
    \caption{\textbf{The dissociation appears only along PC1, and soft attention at small gaps limits the order patch.} \textbf{(a)} Accuracy of each patch (rows) against each target (columns), scored on positions where that target differs from the unpatched running max. The partial patches preferentially match their respective targets; the whole-vector patch also matches the readout target because the whole-vector and readout targets coincide on the latter's changed positions. \textbf{(b)} Outputs where the patched token is still the maximum (left) and where a natural token has overtaken $v_{\mathrm{orig}}$ (right). The order patch outputs $v_{\mathrm{orig}}$ on the left and, at 40\% of positions, also on the right, where the unpatched model outputs the natural maximum. The readout patch outputs $v_{\mathrm{new}}$ on the left and the natural maximum on the right. \textbf{(c)} Order-patch success and the fraction of positions where the patched token gets the top attention score, against the gap between $v_{\mathrm{new}}$ and the competing natural maximum: at small gaps the patched token usually wins the score, but attention is too soft for the output to follow. \textbf{(d)} Rank-one patches along PC2 or 100 random directions transfer nothing, and patches on their complements act like the whole vector. These patches are smaller than patches along PC1 (mean size 2.2 and 1.1, against 12.5), because numbers differ mostly along PC1, so this control does not fully separate the direction from the size of the patch. Intervals: 95\% bootstrap over sequences.}
    \label{fig:app-steering-controls}
\end{figure}

\begin{figure}[h]
    \centering
    \includegraphics[width=0.5\linewidth]{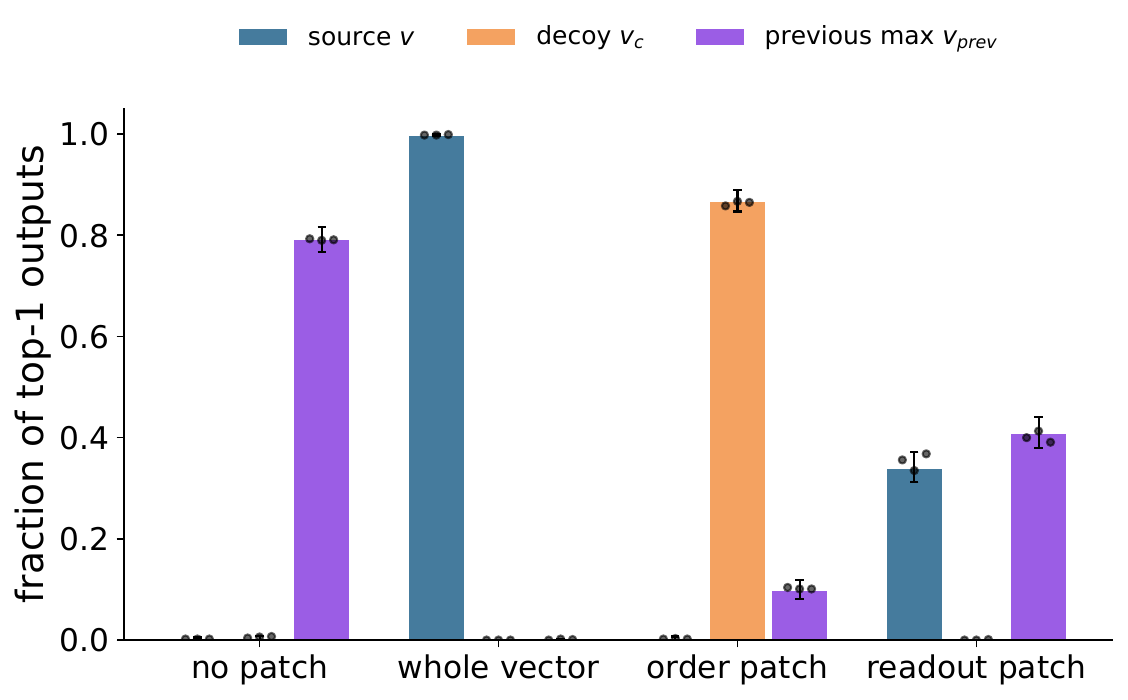}
    \caption{\textbf{With a decoy, the order patch transfers order but keeps the decoy's identity.} Fraction of top-1 outputs equal to the source $v$, the decoy $v_c$ and the previous maximum $v_{\mathrm{prev}}$ (1000 examples). The whole-vector patch outputs $v$ (IIA 0.999). The order patch makes the decoy win the comparison, so the model outputs $v_c$ in 0.87 [0.85, 0.89] of examples. Under the readout patch, the patched token keeps the decoy's order, only 10 below $v_{\mathrm{prev}}$, so the comparison is close and the output splits between $v$ and $v_{\mathrm{prev}}$. The median logit-margin recovery is 1.0 for the whole vector, 0.09 for the order patch and 0.26 for the readout patch. Bars: original run with 95\% bootstrap intervals; dots: three other seeds.}
    \label{fig:app-iia}
\end{figure}

We run two controls for the dissociation in Sec.~\ref{sec:causal}. Fig.~\ref{fig:app-steering-controls} scores each patch against every target, shows the outputs behind these scores, relates the partial order transfer to soft attention, and repeats the patches along PC2 and random directions. Fig.~\ref{fig:app-iia} repeats the experiment with a decoy number, which separates the order and the identity of the patched token at a single readout position.

\subsection{Attention saturation and the interference budget}

\begin{figure}[h]
    \centering
    \includegraphics[width=0.75\linewidth]{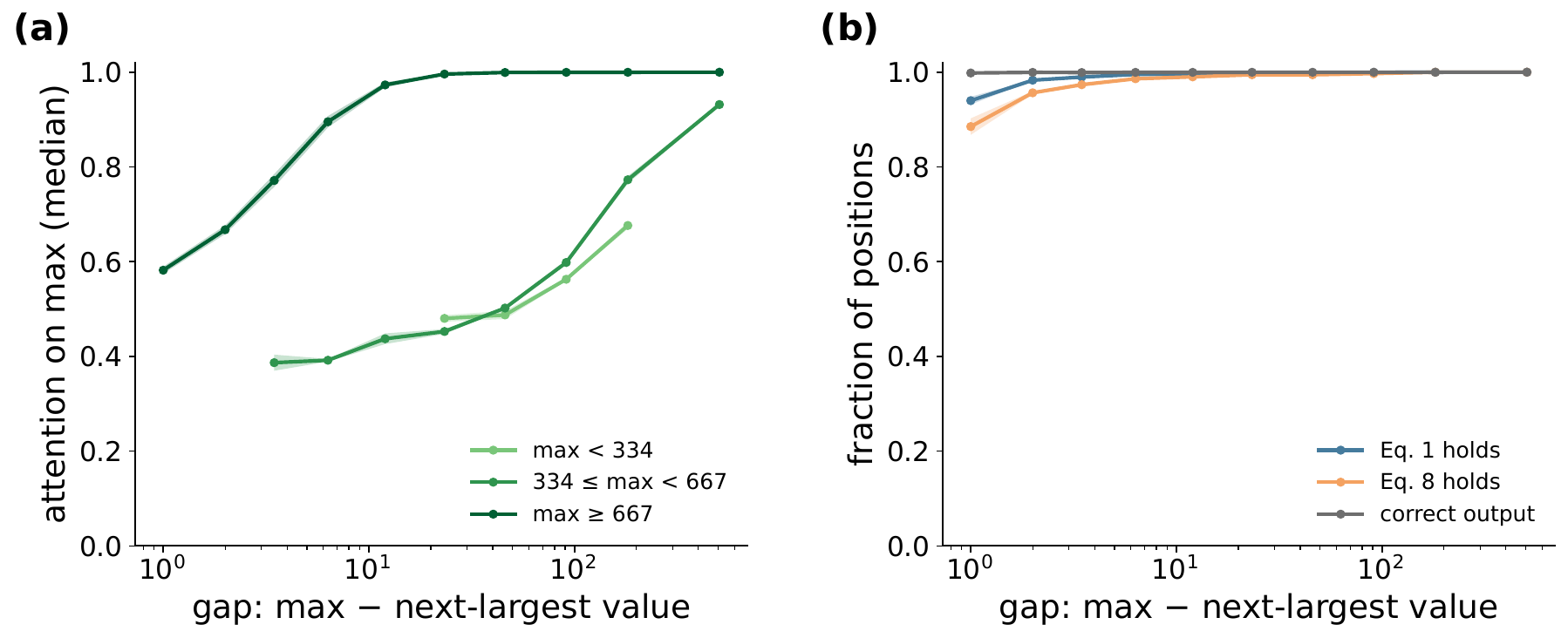}
    \caption{\textbf{Attention is saturated for large maxima but soft for small ones, and the model stays accurate in both regimes.} \textbf{(a)} Median attention on the running maximum against the gap to the next-largest earlier token: above 0.97 for maxima above 667 once the gap is about 9, but between 0.4 and 0.6 for smaller maxima at gaps below 100. \textbf{(b)} Eq.~\eqref{eq:max-condition} and Eq.~\eqref{eq:budget} hold at almost all positions, and accuracy is near 1 even at gaps where they fail more often. 2000 validation sequences; mean and SD over four seeds.}
    \label{fig:app-saturation}
\end{figure}

\begin{figure}[h]
    \centering
    \includegraphics[width=0.75\linewidth]{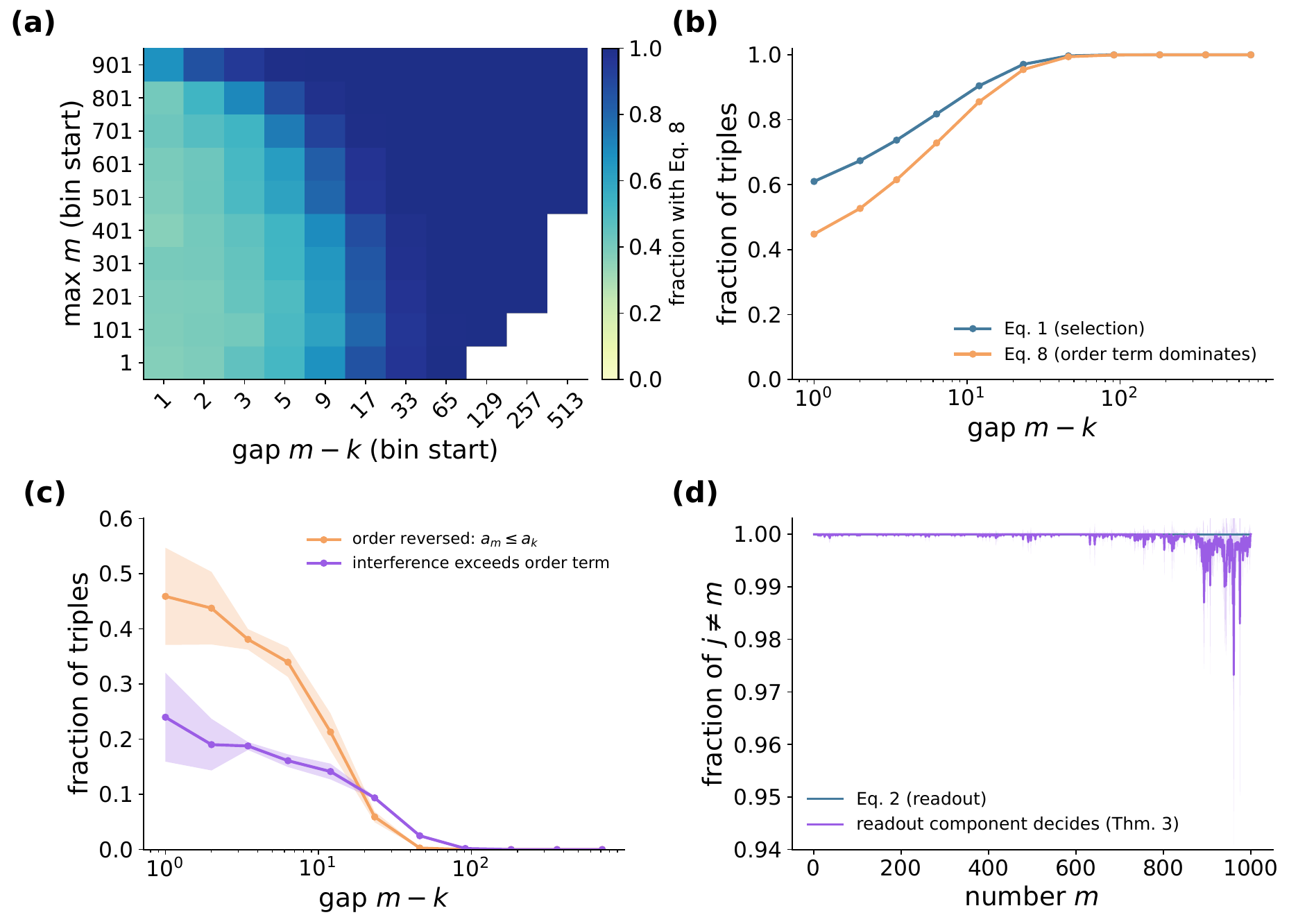}
    \caption{\textbf{The interference budget fails only at small gaps, mostly because PC1 misorders adjacent numbers.} \textbf{(a)} Fraction of triples $(q,m,k)$ that satisfy Eq.~\eqref{eq:budget}, by maximum $m$ and gap $m-k$ (original run): failures sit at small gaps. \textbf{(b)} Eqs.~\eqref{eq:max-condition} and \eqref{eq:budget} against the gap: both exceed 0.95 from a gap of about 20. \textbf{(c)} The two ways Eq.~\eqref{eq:budget} can fail: at small gaps, PC1 puts the pair in the wrong order ($a_m\le a_k$) more often than the interference exceeds a positive order term. \textbf{(d)} For nearly every $m$, the readout condition, Eq.~\eqref{eq:readout-condition}, holds, and the readout component alone decides it, Eq.~\eqref{eq:app-readout-dominance}. Mean and SD over four seeds unless noted.}
    \label{fig:app-interference}
\end{figure}

The theory assumes that attention is saturated and that the order term dominates the interference (Eq.~\eqref{eq:budget}). Fig.~\ref{fig:app-saturation} measures the attention on the running maximum in validation sequences, and Fig.~\ref{fig:app-interference} evaluates Eq.~\eqref{eq:budget} over all vocabulary triples and the readout conditions over all pairs.

\subsection{Two-token inputs: an exact test of the theory}
\label{app:pairs}

\begin{figure}[h]
    \centering
    \includegraphics[width=0.75\linewidth]{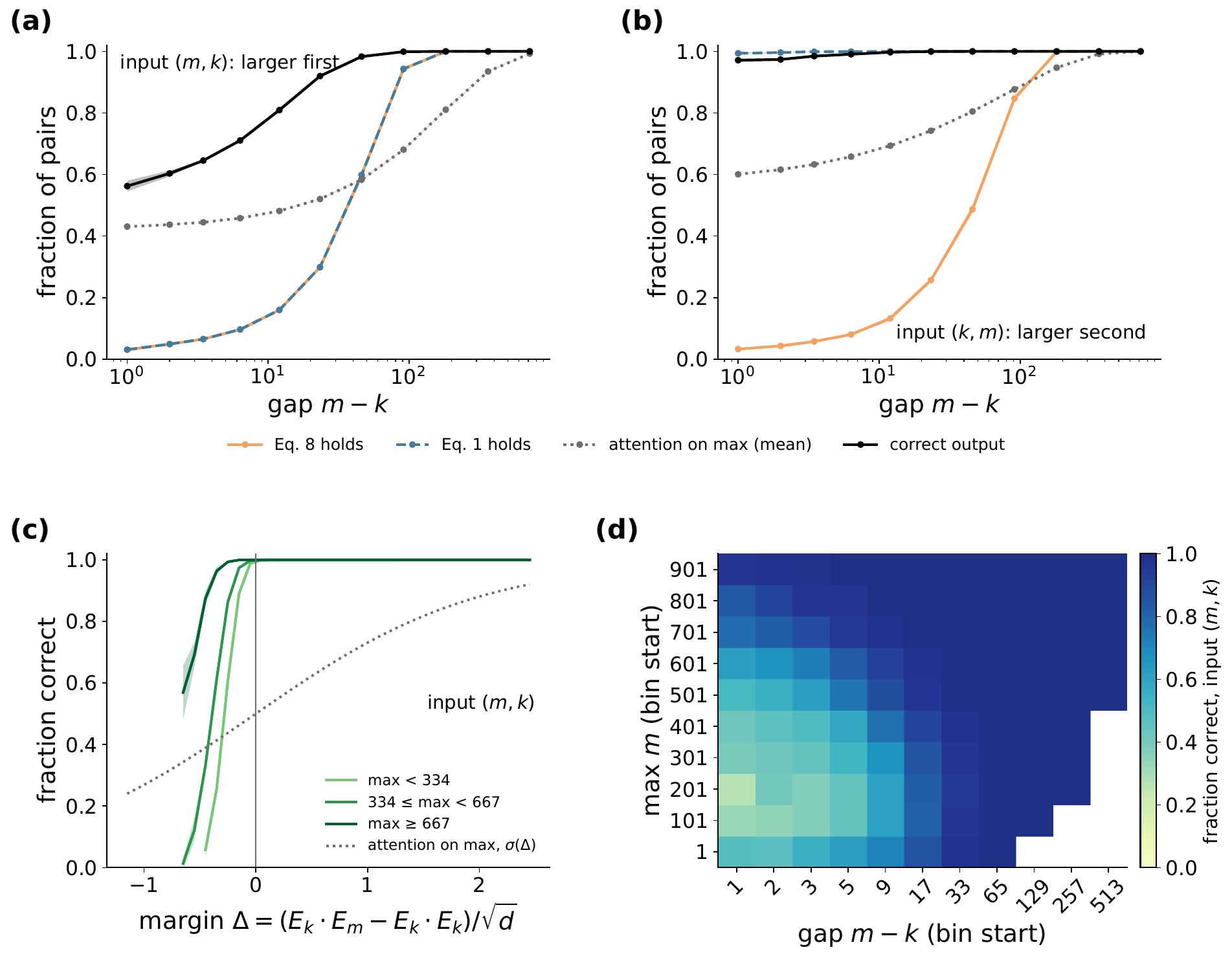}
    \caption{\textbf{The comparison margin predicts errors on two-token inputs.} \textbf{(a)} Larger number first, $(m,k)$: Eqs.~\eqref{eq:max-condition} and \eqref{eq:budget} hold for the same pairs and fail at small gaps, yet accuracy stays far above them, with attention on the maximum below one half. \textbf{(b)} Larger number second, $(k,m)$: Eq.~\eqref{eq:max-condition} always holds and accuracy is near 1, so Eq.~\eqref{eq:budget} is conservative. \textbf{(c)} Accuracy rises from 0 to 1 over a narrow band of negative margins $\Delta$, lower for larger maxima; dotted: attention on the maximum, $\sigma(\Delta)$. \textbf{(d)} Errors concentrate at small gaps and small maxima (larger first, one seed). (a--c): mean and SD over four seeds.}
    \label{fig:pairs}
\end{figure}

\paragraph{Setup.} Every training sequence begins with a pair of numbers, so two-token inputs are in distribution, and they let us test all $499{,}500$ pairs $k<m$ exactly. We run the model on both orders, $(m,k)$ and $(k,m)$, and read the output at the second position. The query $q$ is the second token: $q=k$ for $(m,k)$ and $q=m$ for $(k,m)$. For each pair, we compute the order term $C=a_q(a_m-a_k)$, the interference $I=\r_q\cdot(\r_m-\r_k)$, and the margin $\Delta=(C+I)/\sqrt d$. With two tokens, the attention on the maximum is exactly $\sigma(\Delta)$, where $\sigma$ is the logistic function; we verify this numerically. Input order matters even without positional encoding, because the query is always the last token. With the larger number first, the query $k$ must prefer the earlier $m$ over itself, which is the self-competition behind Theorem~\ref{thm:geometric-conditions}. With the larger number second, the query is the maximum itself. Fig.~\ref{fig:pairs-main} in the main text repeats panels (a) and (c) of Fig.~\ref{fig:pairs}.

\paragraph{The margin predicts errors.} The model is almost perfect when the larger number comes second (accuracy 0.9997) and makes its errors when the larger number comes first (accuracy 0.988; Fig.~\ref{fig:pairs}a,b). These errors concentrate at small gaps and small numbers. For adjacent numbers with the larger first, accuracy is $0.56\pm0.02$, ranging from 0.36 for maxima below 334 to 0.85 for maxima above 667 (Fig.~\ref{fig:pairs}d), and 99\% of errors output the smaller number. The margin predicts these errors slightly better than the gap does: the area under the ROC curve \citep{fawcett2006roc} is 0.990 for $\Delta$ and 0.993 for the order term alone, against 0.979 for the gap $m-k$ (Table~\ref{tab:pairs}).

\paragraph{The readout component biases attention toward the query.} With the larger number first, Eqs.~\eqref{eq:budget} and \eqref{eq:max-condition} hold for the same fraction of pairs (0.919). The sign of the interference explains this (App.~\ref{app:decomposition}): when the query competes with itself, the interference is negative whenever $\cos\theta(\r_q,\r_m)<\|\r_q\|_2/\|\r_m\|_2$. The readout components of different numbers are nearly orthogonal (median cosine 0.12 to 0.14 across seeds), so the interference is negative in over 99\% of pairs, and Eq.~\eqref{eq:budget} then holds exactly when Eq.~\eqref{eq:max-condition} does. With the larger number second, the interference is positive for every pair, so the readout component helps the comparison and Eq.~\eqref{eq:budget} is conservative (0.90 against 1.00).

\paragraph{The model is accurate beyond the saturated regime.} For adjacent numbers with the larger first, Eq.~\eqref{eq:max-condition} holds for only 3\% of pairs, yet 56\% are answered correctly. Accuracy reaches 50\% at negative margins, $\Delta\approx-0.26$, $-0.36$ and $-0.60$ for small, middle and large maxima, where the attention on the maximum is only 0.44, 0.41 and 0.35 (Fig.~\ref{fig:pairs}c). The readout therefore decodes a mixture of the two embeddings in favor of the larger number, and does so more readily for large numbers.

\begin{table}[h]
\centering
\small
\begin{tabular}{lcc}
\toprule
Quantity & Larger first $(m,k)$ & Larger second $(k,m)$ \\
\midrule
Accuracy & $0.9879\pm0.0003$ & $0.9997\pm0.0000$ \\
Accuracy, gap 1 & $0.562\pm0.017$ & $0.971\pm0.006$ \\
Accuracy, gap 1, max $<334$ & $0.362\pm0.028$ & $0.933\pm0.010$ \\
Accuracy, gap 1, max $\ge667$ & $0.849\pm0.009$ & $1.000\pm0.000$ \\
Eq.~\eqref{eq:max-condition} holds & $0.919\pm0.001$ & $1.000\pm0.000$ \\
Eq.~\eqref{eq:budget} holds & $0.919\pm0.001$ & $0.899\pm0.002$ \\
Accuracy when Eq.~\eqref{eq:max-condition} fails & $0.851\pm0.007$ & $0.64\pm0.12$ \\
AUC: margin $\Delta$ & $0.9898\pm0.0003$ & $0.9985\pm0.0001$ \\
AUC: order term $C$ & $0.9928\pm0.0005$ & $0.9961\pm0.0007$ \\
AUC: gap $m-k$ & $0.9790\pm0.0007$ & $0.9906\pm0.0009$ \\
\bottomrule
\end{tabular}
\caption{Two-token inputs: all 499{,}500 pairs $k<m$ in each order. Mean $\pm$ standard deviation over four seeds. AUC: area under the ROC curve for predicting a correct output.}
\label{tab:pairs}
\end{table}

\subsection{Translation symmetry and training dynamics}

\begin{figure}[h]
    \centering
    \includegraphics[width=0.9\linewidth]{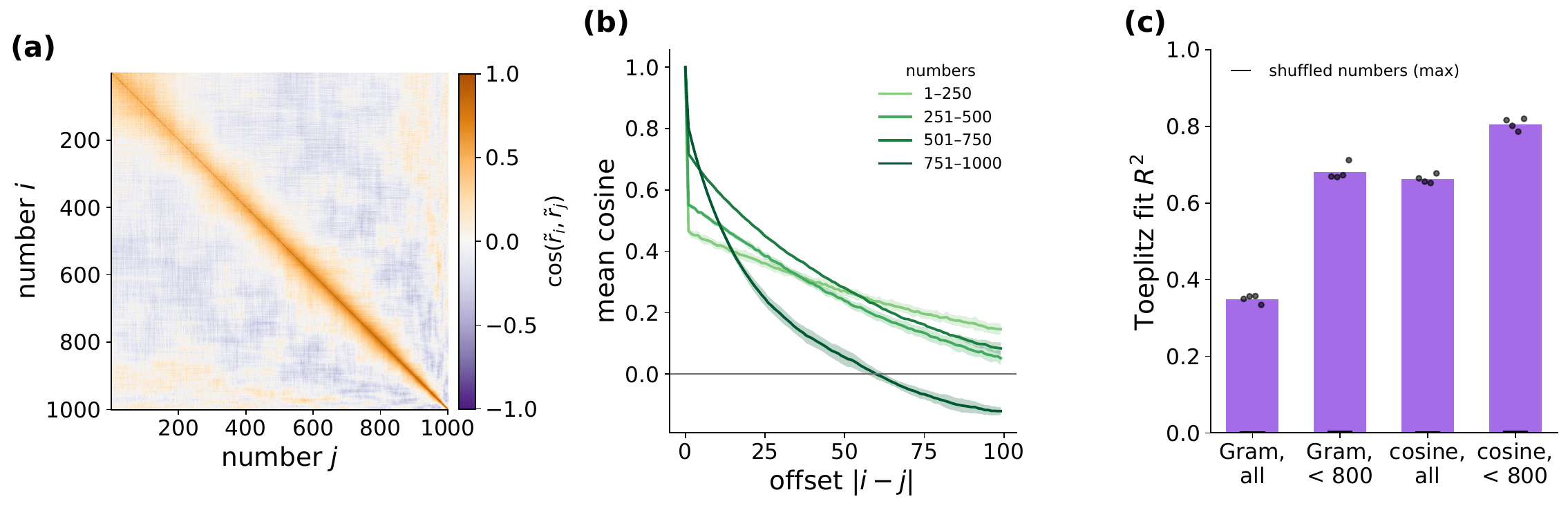}
    \caption{\textbf{Similarities between readout components are close to translation invariant for numbers below about 800.} \textbf{(a)} The cosine similarity between centered readout components $\widetilde\r_i$ and $\widetilde\r_j$ is concentrated near the diagonal (original run). \textbf{(b)} Mean cosine against the offset $|i-j|$, within four blocks of 250 numbers: the blocks up to 750 are similar, and the block above 750 decays faster. \textbf{(c)} A Toeplitz fit $G_{ij}\approx g(|i-j|)$ explains most of the cosine matrix for numbers below 800, and almost nothing after the number labels are shuffled (black lines). Bars: mean over four seeds; dots: seeds.}
    \label{fig:app-toeplitz}
\end{figure}

\begin{figure}[h]
    \centering
    \includegraphics[width=0.75\linewidth]{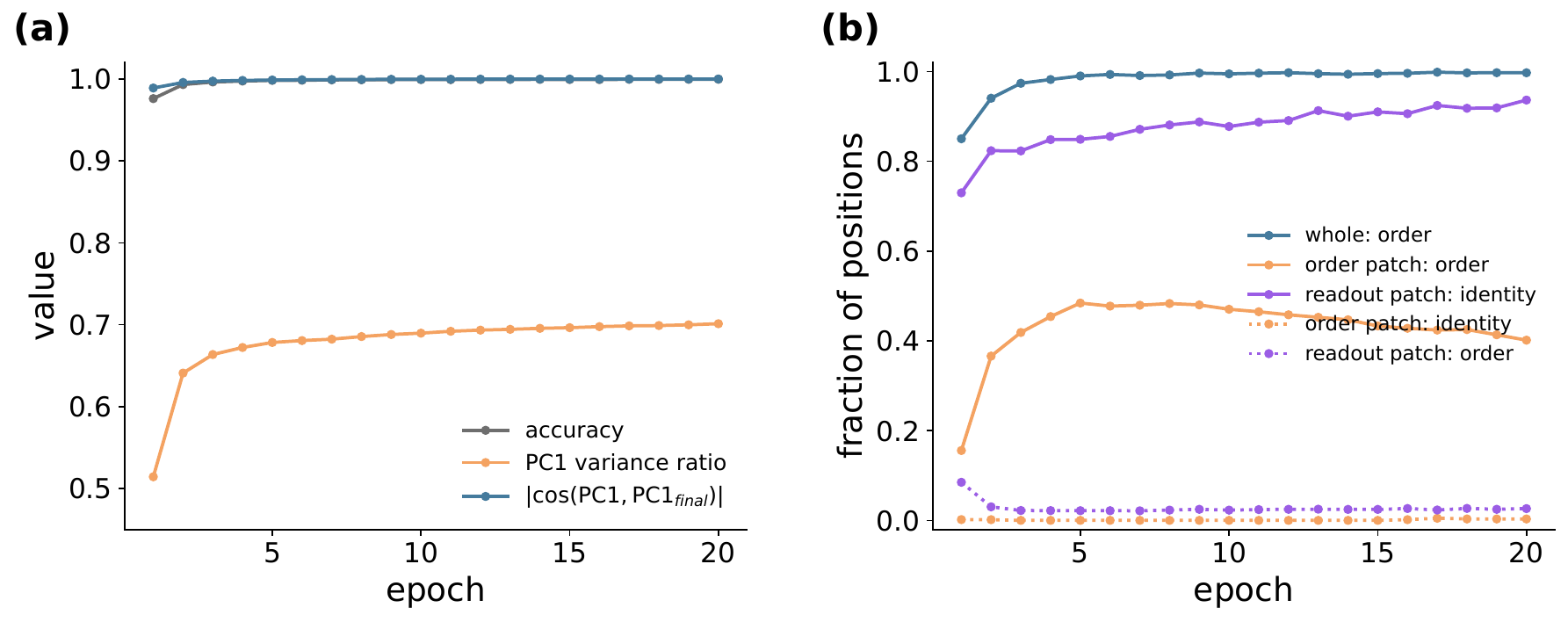}
    \caption{\textbf{The decomposition forms early in training.} \textbf{(a)} Accuracy is above 0.97 after one epoch, and PC1 keeps its final direction from the first epoch on ($|\cos|$ with the final PC1), while its share of the variance grows from 0.51 to 0.70. \textbf{(b)} Order and identity transferred by each patch over training: the identity effect of the readout patch grows steadily, and the order effect of the order patch peaks at epoch 5. Original run.}
    \label{fig:app-epochs}
\end{figure}

Prop.~\ref{prop:translation-symmetry} predicts swirls if the similarities between readout components depend only on the distance between numbers. Fig.~\ref{fig:app-toeplitz} tests this with a Toeplitz fit, and Fig.~\ref{fig:app-epochs} follows the geometry and the effects of the patches over training.

\clearpage
\section{What Determines the Geometry? Architectural Variants}
\label{app:ablations}

The theory links the order direction to comparison through dot-product attention, and it links the need for a readout component to the bias-free readout. We test both links with two variants that each change one component of the model (Table~\ref{tab:variants}). The distance-decoder run uses a smaller dataset and more epochs, so that comparison is not fully matched.

\begin{table}[h]
\centering
\small
\setlength{\tabcolsep}{3.5pt}
\begin{tabular}{llllll}
\toprule
Variant & Attention score & Decoder & Sequences & Epochs & Accuracy \\
\midrule
Baseline & $\bm h_t\cdot\bm h_{t'}/\sqrt d$ & linear & 500{,}000 & 20 & 0.9996 \\
Distance attention & $-\|\bm h_t-\bm h_{t'}\|_2^2/d$ & linear & 500{,}000 & 20 & 0.963 \\
Distance decoder & $\bm h_t\cdot\bm h_{t'}/\sqrt d$ & $\tfrac{2}{\sqrt d}\bm z_t\cdot\U_j-\|\bm z_t\|_2^2-\|\U_j\|_2^2$ & 50{,}000 & 50 & 0.990 \\
\bottomrule
\end{tabular}
\caption{The three models compared. Accuracy is top-1 accuracy over all positions of the full validation split of each run.}
\label{tab:variants}
\end{table}

\begin{figure}[t]
    \centering
    \includegraphics[width=0.85\linewidth]{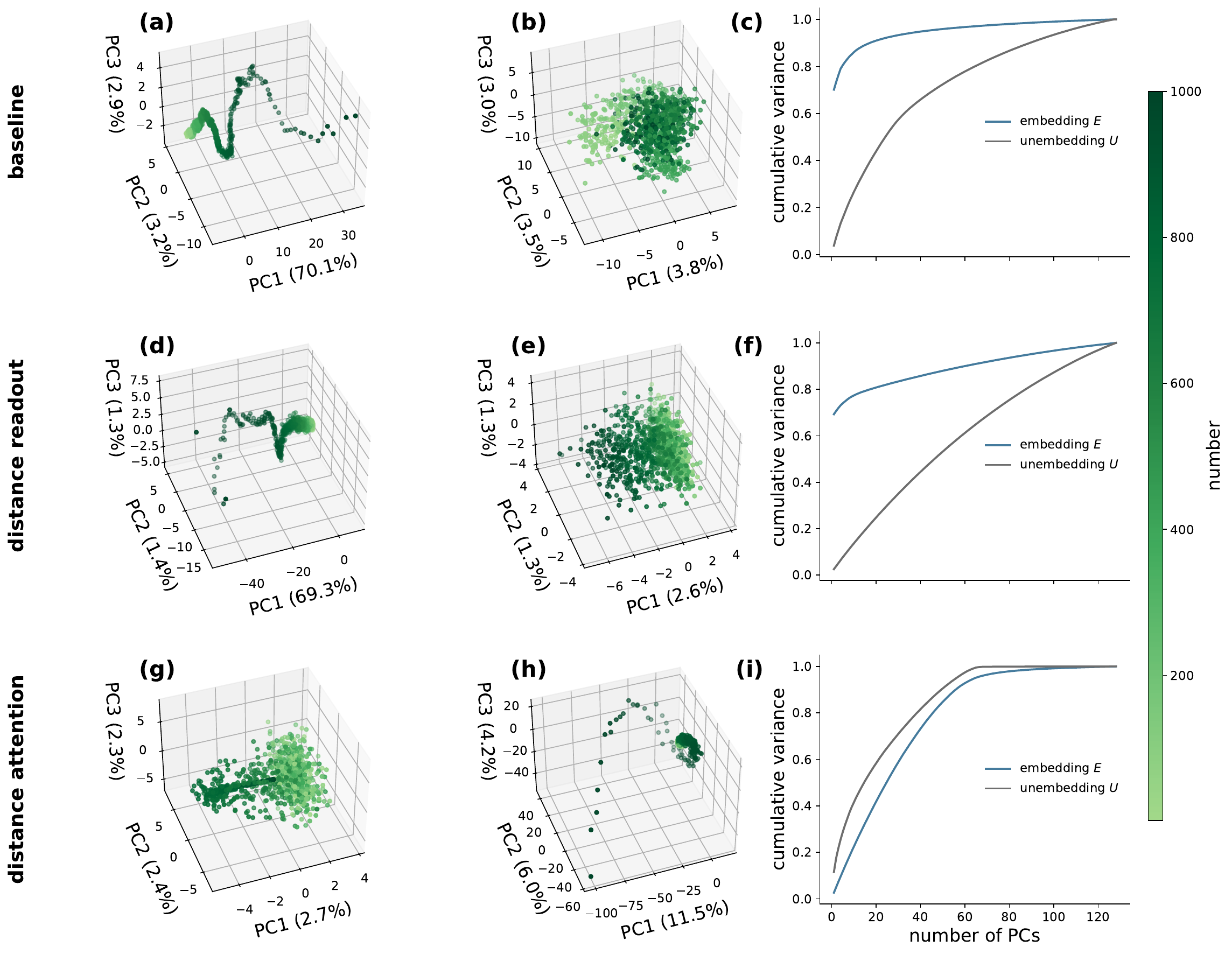}
    \caption{\textbf{Distance-based attention removes the order direction; a distance-based readout does not.} Rows: baseline, distance-based readout, distance-based attention. \textbf{(a, d, g)} Top three PCs of $\E$: the baseline and the distance-readout model share a stretched, ordered manifold, while the distance-attention model has no ordered structure. \textbf{(b, e, h)} Top three PCs of $\U$. \textbf{(c, f, i)} Cumulative variance: one component dominates $\E$ in the first two models, while the distance-attention model spreads its variance over about 60 components. Color: number.}
    \label{fig:ablation-main}
\end{figure}

\begin{figure}[t]
    \centering
    \includegraphics[width=\linewidth]{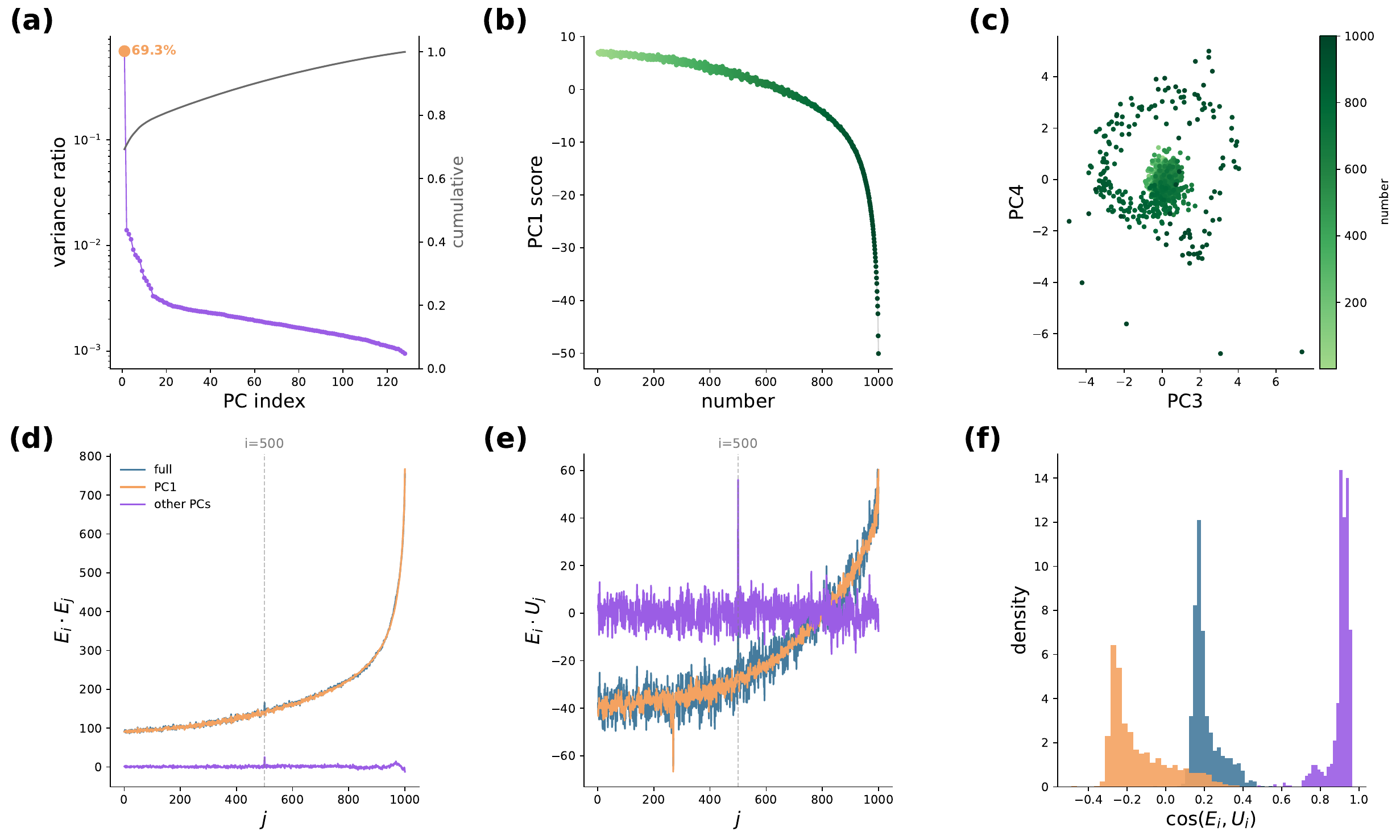}
    \caption{\textbf{A distance-based readout keeps the decomposition into order and identity.} \textbf{(a)} PC1 explains 69.3\% of the variance of $\E$. \textbf{(b)} PC1 changes monotonically with number (its sign is arbitrary). \textbf{(c)} The readout component (other PCs) forms swirls. \textbf{(d)} The order component (PC1) carries almost all of $\E_i\cdot\E_j$. \textbf{(e)} The readout component carries the peak of $\E_i\cdot\U_j$ at $j=i$. \textbf{(f)} The readout component of $\E_i$ is strongly aligned with $\U_i$ (cosine near 0.9), while its order component is anti-aligned.}
    \label{fig:info-identity-distdecoder}
\end{figure}

\paragraph{A distance-based readout keeps the decomposition.} A readout with logits $\tfrac{2}{\sqrt d}\,\bm z_t\cdot\U_j-\|\U_j\|_2^2-\|\bm z_t\|_2^2$, is equivalent, up to a class-independent additive term, to the negative squared distance $-\|\bm z_t/\sqrt d-\U_j\|_2^2$. Equivalently, it is an affine readout with class bias $-\|\U_j\|_2^2$, since $-\|\bm z_t\|_2^2$ is shared across classes. With such a readout, numbers on a ray can in principle be decoded by the unembedding nearest to $\bm z_t/\sqrt d$, so Theorem~\ref{thm:norm-efficient-selection} no longer rules out a ray, while comparison through dot-product attention is unchanged. The learned geometry is nevertheless the same as in the baseline (Figs.~\ref{fig:ablation-main}d--f and \ref{fig:info-identity-distdecoder}). Embedding norms grow with number, PC1 explains 69.3\% of the variance, changes monotonically with number and carries $\E_i\cdot\E_j$, and the other PCs form swirls and carry the alignment with $\U_i$. In this run, the readout component does not depend on the bias-free readout, which Theorem~\ref{thm:norm-efficient-selection} alone cannot explain. The variant also serves as a readout with class-dependent biases, although its bias $-\|\U_j\|_2^2$ is tied to the unembedding.

\paragraph{Distance-based attention removes the order direction.} With attention scores $-\|\bm h_t-\bm h_{t'}\|_2^2/d$, every token is at distance zero from itself, so a query always gives itself the highest score. Eq.~\eqref{eq:max-condition} then fails whenever the query is smaller than the maximum, saturated retrieval of a larger earlier number becomes impossible, and none of the conditions of Sec.~\ref{sec:theory} apply. This model reaches 96.3\% accuracy through a different mechanism, which we do not analyze. Its embeddings have no dominant direction (Fig.~\ref{fig:ablation-main}g--i): PC1 explains only 2.7\% of the variance, does not change monotonically with number and does not carry $\E_i\cdot\E_j$, and about 60 components are needed to explain nearly all of the variance. Its unembeddings, in contrast, are more structured than its embeddings (Fig.~\ref{fig:ablation-main}h).

Changing how numbers are read out thus leaves the decomposition intact, while changing how they are compared removes the order direction.

\end{document}